\documentclass{article}
\usepackage{ijcai25}

\usepackage{times}
\usepackage{soul}
\usepackage{url}
\usepackage[hidelinks]{hyperref}
\usepackage[utf8]{inputenc}
\usepackage[small]{caption}
\usepackage{graphicx}
\usepackage{amsmath}
\usepackage{amsthm}
\usepackage{booktabs}
\usepackage{algorithm}
\usepackage{algorithmic}
\usepackage[switch]{lineno}

\title{Online 3D Gaussian Splatting Modeling with Novel View Selection}

\author{
Byeonggwon Lee$^1$
\and
Junkyu Park$^1$\and
Khang Truong Giang$^{2}{}^{*}$\And
Soohwan Song$^1$\footnote{corresponding authors}\\
\affiliations
$^1$Department of Computer Science and Artificial Intelligence, Dongguk University, Seoul, Korea\\
$^2$42dot, Seongnam-si, Gyeonggi-do, Korea\\
\emails
\{lbg030, dannypk99\}@dgu.ac.kr,
khangtg.hust.it@gmail.com,
songsh@dongguk.edu
}

\usepackage{multirow} 
\usepackage{graphicx} 
\usepackage{amsmath} 
\usepackage{booktabs} 
\usepackage{tabularx}
\usepackage{enumitem}
\usepackage{makecell}
\usepackage{verbatim}
\usepackage{adjustbox}
\usepackage{subcaption}
\usepackage{xcolor}

\usepackage{arydshln}
\usepackage{tikz}

\usetikzlibrary{calc}
\begin{document}

\maketitle

\begin{abstract}
This study addresses the challenge of generating online 3D Gaussian Splatting (3DGS) models from RGB-only frames.
Previous studies have employed dense SLAM techniques to estimate 3D scenes from keyframes for 3DGS model construction. However, these methods are limited by their reliance solely on keyframes, which are insufficient to capture an entire scene, resulting in incomplete reconstructions. Moreover, building a generalizable model requires incorporating frames from diverse viewpoints to achieve broader scene coverage. However, online processing restricts the use of many frames or extensive training iterations. Therefore, we propose a novel method for high-quality 3DGS modeling that improves model completeness through adaptive view selection. By analyzing reconstruction quality online, our approach selects optimal non-keyframes for additional training. By integrating both keyframes and selected non-keyframes, the method refines incomplete regions from diverse viewpoints, significantly enhancing completeness. We also present a framework that incorporates an online multi-view stereo approach, ensuring consistency in 3D information throughout the 3DGS modeling process. Experimental results demonstrate that our method outperforms state-of-the-art methods, delivering exceptional performance in complex outdoor scenes.
\end{abstract}

\section{Introduction}
Online 3D modeling from an image stream is a critical challenge in robotics, virtual reality, augmented reality, and related fields. Visual \textit{Simultaneous Localization And Mapping} (SLAM) has emerged as a key solution to this challenge. In particular, dense SLAM techniques \cite{newcombe2011dtam} \cite{teed2021droid} focus on real-time depth estimation to enable detailed 3D scene reconstruction. Recently, dense SLAM \cite{rosinol2023nerf} \cite{zhang2023go} \cite{zhu2024nicer} has been further developed to incorporate neural scene representations \cite{mildenhall2021nerf} \cite{kerbl20233d} \cite{RePaint-NeRF}, enhancing high-quality rendering and realistic scene synthesis. Among these advancements, \textit{3D Gaussian Splatting} (3DGS) \cite{kerbl20233d} has attracted significant attention for its efficient 3D scene representation and real-time rendering capabilities.

\begin{figure}[t]
    \centering
    \includegraphics[width=0.95\linewidth]{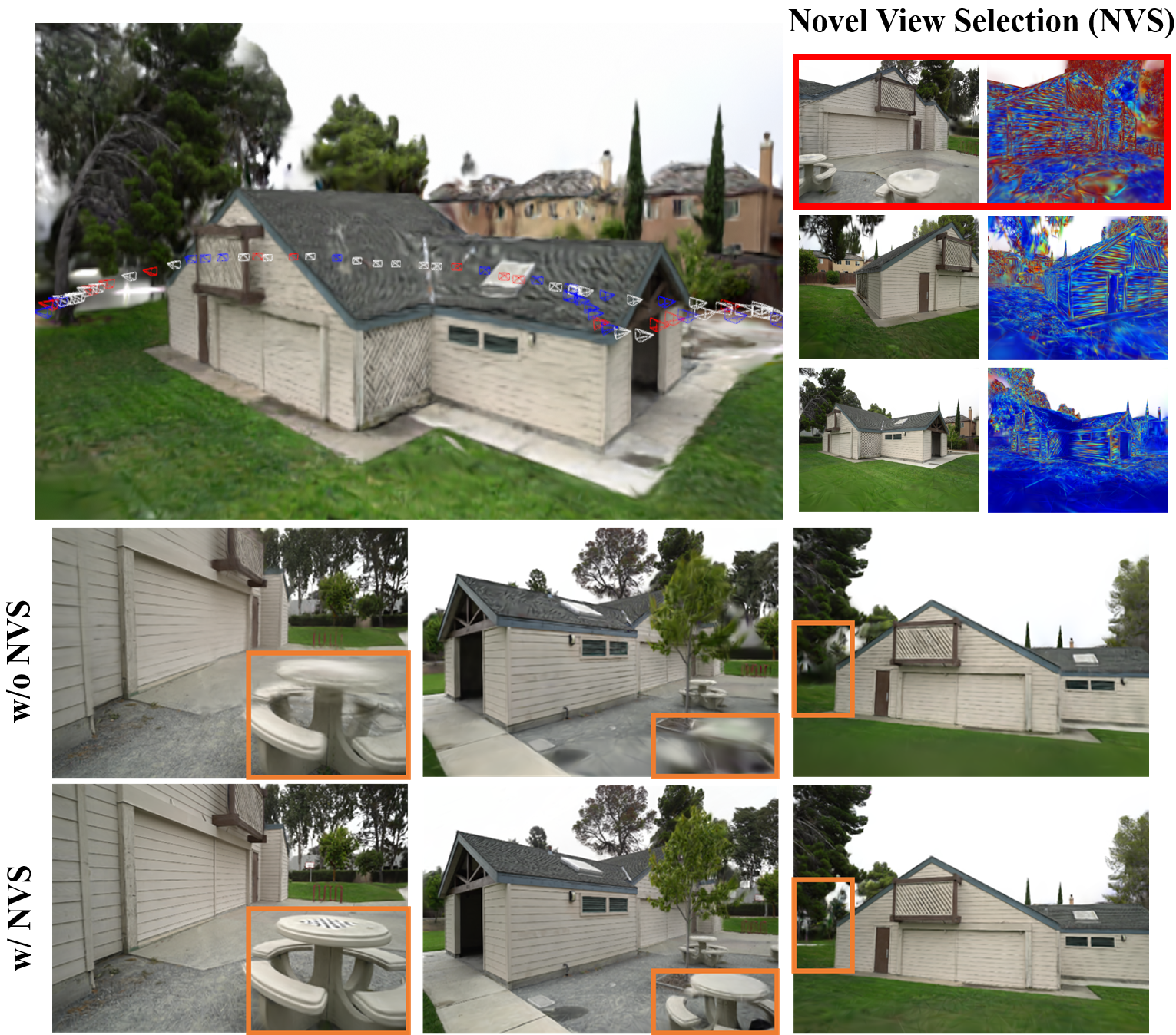}
    \caption{The main idea of our novel view selection for online 3DGS modeling. (\textit{upper}) Our method selects non-keyframes that observe the most uncertain Gaussians (red regions in the uncertainty map) for additional training. In the figure, blue views represent keyframes, red views represent selected non-keyframes, and white views represent the remaining non-keyframes. (\textit{lower}) This approach enhances the completeness of 3DGS, yielding high-quality renderings.}
    \label{fig1: overview}
\end{figure}

In this study, we tackle the challenge of generating online 3DGS models from RGB-only frames. Previous works \cite{matsuki2024gaussian} \cite{huang2024photo} \cite{sandstrom2024splat} have focused mainly on dense SLAM frameworks that utilize 3DGS as a map representation. These methods achieve real-time depth estimation through techniques such as optical flow \cite{teed2020raft}, depth prediction \cite{eftekhar2021omnidata}, or motion stereo \cite{song2021view}, integrating the estimated depths directly into the 3DGS model. While these approaches have shown promising results, they still face notable limitations that require further attention.

First, many studies rely on sparse or low-resolution depths to ensure real-time processing, which inherently limits the quality of the final model. Notably, some approaches \cite{zhu2024nicer} \cite{sandstrom2024splat} employ depth prediction networks \cite{eftekhar2021omnidata} that infer depth directly from images but often generate ambiguous estimates. Second, existing methods \cite{huang2024photo} \cite{sandstrom2024splat} \cite{lee2024mvs} store depth information only for keyframes, restricting 3DGS model training to these frames. This limitation can lead to incomplete reconstruction not only of unobserved areas but also of covered 3D scenes due to insufficient keyframe coverage. Lastly, online 3D modeling must operate within a constrained timeframe, which restricts the number of image frames and training iterations that can be processed. To achieve the best results within these constraints, it is essential to identify and prioritize the frames that most effectively enhance the performance of 3DGS during training.

To address these challenges, we propose a novel method that adaptively improves model completeness through novel view selection. Our approach identifies an optimal set of non-keyframes for additional training by analyzing the online reconstruction quality of the 3DGS model. Specifically, we define the uncertainty of Gaussian primitives based on their shapes and the gradients of their positions. This uncertainty metric is then used to calculate the information gain from a given viewpoint, enabling the selection of optimal views. By incorporating both keyframes and selected non-keyframes, our method refines incomplete regions from diverse viewpoints, resulting in improved completeness (see Fig.~1).

Additionally, we present an efficient framework for online 3DGS mapping that integrates an independent frontend and backend for streamlined processing. The frontend leverages a state-of-the-art \textit{multi-view stereo} (MVS) network \cite{cao2022mvsformer} to generate accurate depths from sequential input images. Operating in parallel, the backend focuses on optimizing 3DGS models, allowing sufficient time for effective parameter refinement. To maintain consistency between the frontend and backend, the framework continuously applies \textit{global bundle adjustment} (GBA) and Gaussian deformation. This approach achieves superior rendering accuracy compared to state-of-the-art methods, demonstrating exceptional performance in handling complex outdoor environments.

The main contributions are summarized as follows:

\begin{itemize}[leftmargin=0.5cm]
    
    \item We propose an online 3DGS modeling method that leverages novel view selection. This approach selects optimal views from non-keyframes for additional training, significantly improving model completeness.

    \item We define the uncertainty of Gaussians using their shapes and positional gradients. This approach allows for more effective view selection compared to other uncertainty metrics \cite{jiang2025fisherrf} \cite{jin2024gs}.

    \item We present a 3DGS mapping framework that utilizes an online MVS approach. This framework ensures consistent 3D information throughout the entire process.

    \item The proposed method was evaluated using two benchmarks for indoor scenes \cite{sturm2012benchmark} \cite{straub2019replica}. To highlight its generalization capability, we also extended the evaluation to include challenging outdoor scenarios \cite{song2021view} \cite{knapitsch2017tanks}.
\end{itemize}

\section{Related Works}
\subsection{Monocular Dense SLAM}
Monocular SLAM \cite{mur2015orb} \cite{forster2014svo} traditionally generates sparse feature maps from monocular images and uses them to estimate camera poses. These methods have mainly focused on achieving precise pose estimation and have achieved significant success. The focus has shifted to reconstructing detailed 3D scenes, fueling the development of dense SLAMs \cite{engel2017direct}. 

With deep learning achieving state-of-the-art performance in optical flow \cite{teed2020raft} and depth estimation \cite{eftekhar2021omnidata} \cite{yao2018mvsnet}, many studies also have investigated the incorporation of deep learning modules into dense SLAM systems \cite{bloesch2018codeslam,koestler2022tandem,teed2021droid}. DROID-SLAM \cite{teed2021droid} combines an optical flow network with GBA to achieve precise trajectory estimation and dense 3D modeling.


\begin{figure*}[t]  
    \centering
    \includegraphics[width=0.99\textwidth]{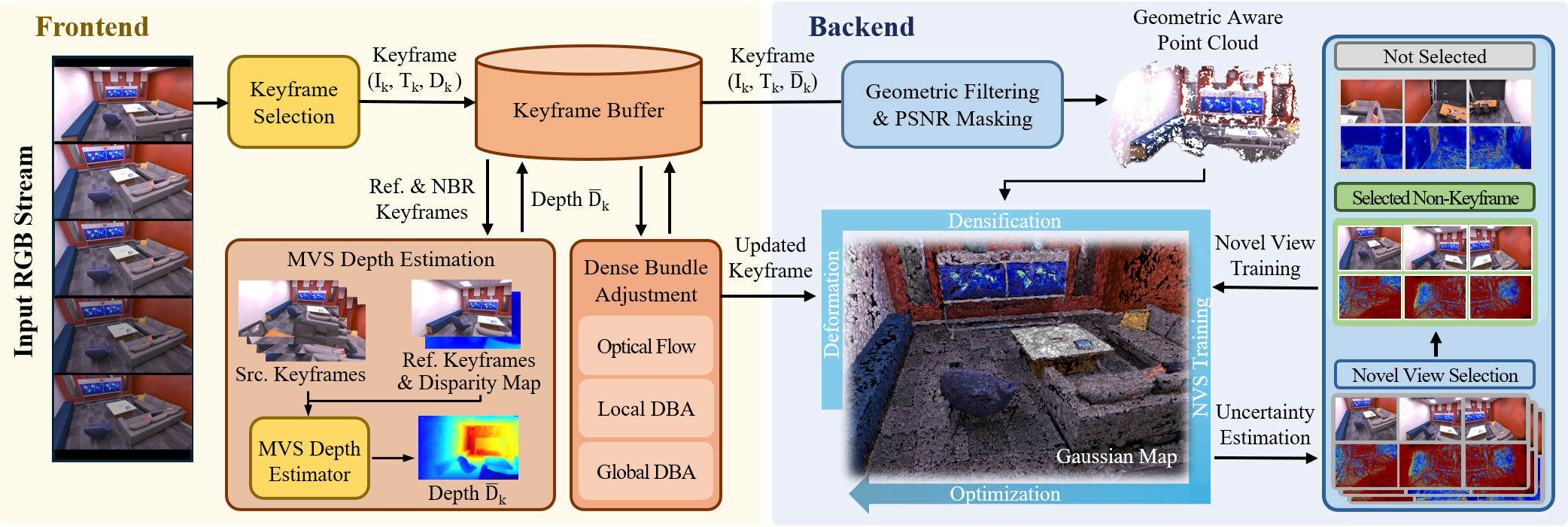}  
    \caption{System overview. The frontend tracks camera poses and estimates depth maps for keyframes using MVS networks, with the disparity map derived from tracking module used to initialize the network. The backend generates new Gaussians from keyframes and continuously optimizes the 3DGS model. To further improve performance, our system periodically performs GBA and conducts additional training on non-keyframes selected through NVS.
    }
    \label{fig2: System_framework}
\end{figure*}

\subsection{Differentiable Rendering SLAM}
Differentiable volumetric rendering, popularized by NeRF \cite{mildenhall2021nerf}, has recently emerged as a key technique for achieving photorealistic scenes and has been applied to dense SLAM. GO-SLAM \cite{zhang2023go} and NeRF-SLAM \cite{rosinol2023nerf} integrate low-resolution depth maps estimated from DROID-SLAM \cite{teed2021droid} into NeRFs. NICER-SLAM \cite{zhu2024nicer} and GLORIE-SLAM \cite{zhang2024glorie} leverage depth prediction networks to generate high-resolution depth maps, which are further utilized to improve NeRFs.

Recently, 3DGS \cite{kerbl20233d,ma2024fastscene}  has gained attention as a promising solution for 3D modeling, offering faster rendering speeds compared to NeRF. By representing 3D space with Gaussian primitives, 3DGS achieves efficient rendering through rasterization and spatial optimization, making it highly suitable for dense SLAM. Photo-SLAM \cite{huang2024photo} and MGSO \cite{hu2024mgso} generate Gaussian points from sparse feature points extracted using ORB-SLAM \cite{mur2015orb} and Direct Sparse Odometry \cite{engel2017direct}, respectively. Splat-SLAM \cite{sandstrom2024splat} further enhances 3DGS mapping by combining proxy depth maps with depth prediction results. It employs deformable Gaussians and globally optimized optical flow tracking, resulting in improved accuracy.

Existing methods \cite{kerbl20233d} \cite{matsuki2024gaussian} \cite{huang2024photo} \cite{hu2024mgso} \cite{sandstrom2024splat} suffer from reduced quality due to their reliance on sparse or imprecise depth data, with depth prediction performing poorly in untrained environments like outdoor scenes. To address this limitation, MVS-GS \cite{lee2024mvs} leverages an online MVS approach \cite{song2021view} to estimate high-quality depth maps, enabling accurate Gaussian initialization. Building on the MVS-GS framework, our method further improves 3DGS mapping by incorporating iterative GBA with Gaussian deformation. Additionally, we propose a novel view selection strategy that prioritizes non-keyframes with high information gain, integrating them into the training process. These enhancements significantly improve the completeness and quality of the resulting 3DGS model.

\subsection{Next-Best-View Planning}
The next-best-view (NBV) problem \cite{scott2003view} focuses on identifying optimal viewpoints to gather specific information. This process aims to maximize information gain, typically defined based on uncertainty, which varies depending on the type of 3D modeling.

With advancements in NeRF, many NBV-related studies have emerged. These approaches estimate uncertainty by analyzing factors such as occupancy probability \cite{yan2023active}, color similarity \cite{ran2023neurar}, or implicit neural representations \cite{feng2024naruto} derived from NeRF models. However, NBV research specifically tailored to 3DGS remains limited. FisherRF \cite{jiang2025fisherrf} quantifies information gain using Fisher information in Gaussian parameters, but it relies on computationally intensive gradient calculations.
GS-Planner \cite{jin2024gs} evaluates completeness by comparing rendered color and depth values with their captured ones, using this measure to estimate information gain.

These methods \cite{jiang2025fisherrf,jin2024gs} measure uncertainty using rendered images; however, they provide inaccurate results when the original image contains noise and are heavily affected by less significant areas, such as the sky. In contrast, the proposed method effectively represents the quality of each Gaussian by leveraging its shape and gradients, allowing for efficient and precise uncertainty estimation.

\section{Proposed Method}
We present a system for online 3DGS modeling using RGB images, designed for high-quality rendering. This system is built on the MVS-GS framework \protect\cite{lee2024mvs}, featuring parallel operation of frontend and backend modules. For the frontend module, we directly adopt the tracking method of \protect\cite{teed2021droid} and the MVS networks from \protect\cite{cao2022mvsformer} to simultaneously estimate camera poses and depth maps from an image stream. Based on the outputs of the frontend, we propose novel techniques to learn a high-fidelity 3DGS scene representation in the backend module. 

\subsection{System Overview}
Fig.~2 depicts an overview of the proposed system. Given an input image stream, the frontend dynamically constructs a frame-graph $G$, where each node $I_k$ represents a selected keyframe, and each edge $(I_k,I_l)$ denotes a co-visibility connection between two keyframes. During the construction of the graph $G$, the frontend tracks the camera pose and estimates the depth map for each inserted keyframe.
To mitigate the effects of pose and scale drift and enhance global geometric consistency, we perform GBA online on the entire graph $G$ every 30 keyframes.
The original tracking system \protect\cite{teed2021droid} performs GBA only at the end of the mapping process. In contrast, we continuously perform GBA to consistently optimize the global consistency of 3D model.

Next, the frontend estimates accurate and reliable depth maps based on the MVS network, MVSFormer \protect\cite{cao2022mvsformer}. Depth prediction methods commonly employed in dense SLAM \protect\cite{ranftl2021vision} \protect\cite{bhat2023zoedepth} suffer from scale ambiguity issues due to the inherent limitations of single-image-based estimation, which can result in reduced accuracy and consistency in depth estimation. In contrast, MVS leverages information from surrounding frames, effectively overcoming the scale ambiguity inherent in single-frame estimation and generating more accurate and consistent depth maps. We compute the coarse depths based on the dense optical flow results estimated during the tracking step and use it to initialize the depth map in the first layer of MVSFormer. This approach improves consistency between the tracking and MVS stages, enhancing the robustness of depth estimation while reducing computational costs.

The backend module consistently trains a 3DGS model for scene representation using the estimated camera pose ${T_k}$ and depth map ${\bar{D}_k}$ of each keyframe ${I_k}$. The module begins by using the depths of some initial keyframes to initialize a 3DGS model. Subsequently, it incrementally integrates new 3D Gaussians from incoming keyframes into the existing model (Section 3.2). Additionally, we perform a \textit{novel view selection} (NVS) strategy to identify a set of non-keyframes that significantly enhance 3DGS quality, using them for further training of the 3DGS model (Section 3.3). The NVS quantifies the uncertainty of Gaussians and determines the non-keyframes that cover the largest number of uncertain Gaussians. Our method uses the selected frames for additional training, significantly improving the completeness of the 3DGS model and the generalizability of its rendering.

\subsection{Online 3DGS Modeling}
\noindent\textbf{Map Construction.}
\noindent For each keyframe, we first transform the depth map $\bar{D}_k$ into a point cloud based on the camera pose $T_k$. We remove noisy points and outliers from the point cloud to obtain reliable and consistent 3D points. This step was implemented using geometric and photometric consistency criteria commonly employed in MVS \protect\cite{cao2022mvsformer}.

The initial point cloud, generated from the first few keyframes, is used to create a set of Gaussian primitives for the 3DGS model. As new keyframes are extracted, the 3DGS model is incrementally updated by incorporating additional Gaussians. Following the approach of \protect\cite{lee2024mvs}, we generate new Gaussians exclusively from unexplored regions with low rendering quality to improve computational efficiency. The unexplored regions are identified by comparing the rendered image $\hat{I_k}$ with the original images $I_k$ using the peak signal-to-noise ratio (PSNR), marking those below a threshold as such regions. Throughout the map construction process, we regularly perform GBA and Gaussian optimization to refine camera poses and update the model parameters.

\noindent\textbf{Gaussian Representation.}
\noindent We utilize the generalized exponential splatting (GES) \protect\cite{hamdi2024ges} to parameterize each Gaussian in a 3DGS model. Specifically, a Gaussian point $g_{i}$ is defined by several attributes: a covariance matrix $\sum_{i} \in R^{3 \times 3}$, a mean  $\mu_{i} \in R^3$, opacity $o_{i} \in [0,1]$, color $c_i \in R^3$, and a shape parameter $\beta_i \in (0,\infty)$. The shape parameter $\beta_{i}$, which is learnable, controls the sharpness of the Gaussian. The GES framework efficiently captures high-frequency details of the scene using fewer primitives, making it well-suited for real-time mapping and rendering. The Gaussian density function in GES is defined as follows:
\begin{equation}
g_i(x) = \exp \left\{ -\frac{1}{2} (x - \mu_i)^T \Sigma_i^{-1} (x - \mu_i) \right\}^2 \beta_i
\tag{1}
\end{equation}

To render an image $\hat{I}_k$ from an input pose $T_k$, the 3D Gaussians are projected onto the image plane. The color $\hat{c}(p)$ of each pixel $p$ is determined by sorting the Gaussian points by depth and performing alpha blending from front to back:  
\begin{equation}
\hat{c}\left(p\right)=\sum_{i\in N}{c_i\alpha_i\prod_{j=1}^{i-1}\left(1-\alpha_j\right)}
\tag{2}
\end{equation}
where $\alpha_{i} = o_{i}g_{i}(x)$ ensures that closer Gaussian points contribute more to the color of the pixels. Similarly, the depth $\hat{D}_{k}$ can be rendered using the same projection approach:
\begin{equation}
\hat{D}\left(p\right)=\sum_{i\in N}{z_i\alpha_i\prod_{j=1}^{i-1}\left(1-\alpha_j\right)}
\tag{3}
\end{equation}
where $z_i$ is the distance to the Gaussian mean along the ray.

\noindent \textbf{Map Deformation.}
\noindent During incrementally building the 3D Gaussian map, the GBA module updates camera poses and disparity map of keyframes in entire history. Therefore, the 3D Gaussian map should be updated accordingly after each GBA update. However, our map is only rigidly deformed based on the updated camera poses instead of using non-rigid deformation as in \protect\cite{sandstrom2024splat}. The 3D data generated by MVS is highly accurate, allowing for the generation of a consistent model with rigid deformation. This approach helps improve the efficiency of the deformation process.
The map deformation process is executed after the GBA and before the Gaussian optimization step. This helps refine the entire Gaussian map to adapt to the updated camera poses.

\subsection{3DGS Optimization with Novel View Selection}

Existing online methods \cite{sandstrom2024splat,huang2024photo,lee2024mvs} learn Gaussian parameters using only tracked keyframes. The keyframes are usually selected by thresholding mean of optical flow between image frames \protect\cite{teed2020raft}. Those keyframes sometimes cannot fully cover the entire scene due to fast motion changes or occlusions. Additionally, it is infeasible to use all frames, including both keyframes and non-keyframes, for optimization. To address this, our NVS selects top-k non-keyframes that cover poorly reconstructed regions, based on estimated information gain.

\noindent \textbf{Uncertainty Estimation.}
\noindent For each non-keyframe $I_{n}$, we extract visible Gaussians and compute an uncertainty score based on their shapes and positional gradients. For each Gaussian $g_{n,i}$ visible in $I_{n}$, we extract the largest eigenvalue of its covariance $\Sigma_{n,i}$. Here, $\Sigma_{n,i}=R_{n,i}S_{n,i}S_{n,i}^TR_{n,i}^T$, where $R_{n,i}$ and $S_{n,i}$ represent the orientation and scale components of Gaussian $g_{n,i}$, respectively. Since $S_{n,i}=diag\left(s_{n,i}\right)$ where $s_{n,i}\in{R}^3$, it follows that $S_{n,i}S_{n,i}^T=diag\left(s_{n,i}^2\right)$. The largest eigenvalue $\lambda_{n,i}$ is then calculated as:
\begin{equation}
    \lambda_{n,i}=\max{\left(s_{n,i}^2\right)}
\tag{6}
\end{equation}
\noindent This eigenvalue $\lambda_{n,i}$ represents the largest size of $g_{n,i}$ in the direction corresponding to the rotation $R_{n,i}$. Incorporating the eigenvalue into the uncertainty formulation helps address the over-reconstruction indicated by large Gaussians \cite{hu2025cg,matsuki2024gaussian}.

The second component of uncertainty is derived from the adaptive density control mechanism in existing 3DGS models \protect\cite{matsuki2024gaussian}\protect\cite{sandstrom2024splat}.
A large positional gradient indicates that the 3DGS model is still undergoing optimization. Therefore, high gradients suggest that the parameters have not yet been fully optimized \cite{rota2024revising} \cite{kerbl20233d}.
Therefore, we adopted the gradient as a factor for measuring the uncertainty of Gaussians.
Let $d\mu_{n,i}\in R^3$ be the positional gradient of Gaussian $g_{n,i}$. We calculate the magnitude of the gradient $A_{n,i}=||d\mu_{n,i}||$ as a measure of uncertainty estimation. This approach helps prioritize regions requiring further optimization.

\begin{figure}[t]
    \centering
    \includegraphics[width=1.000\linewidth]{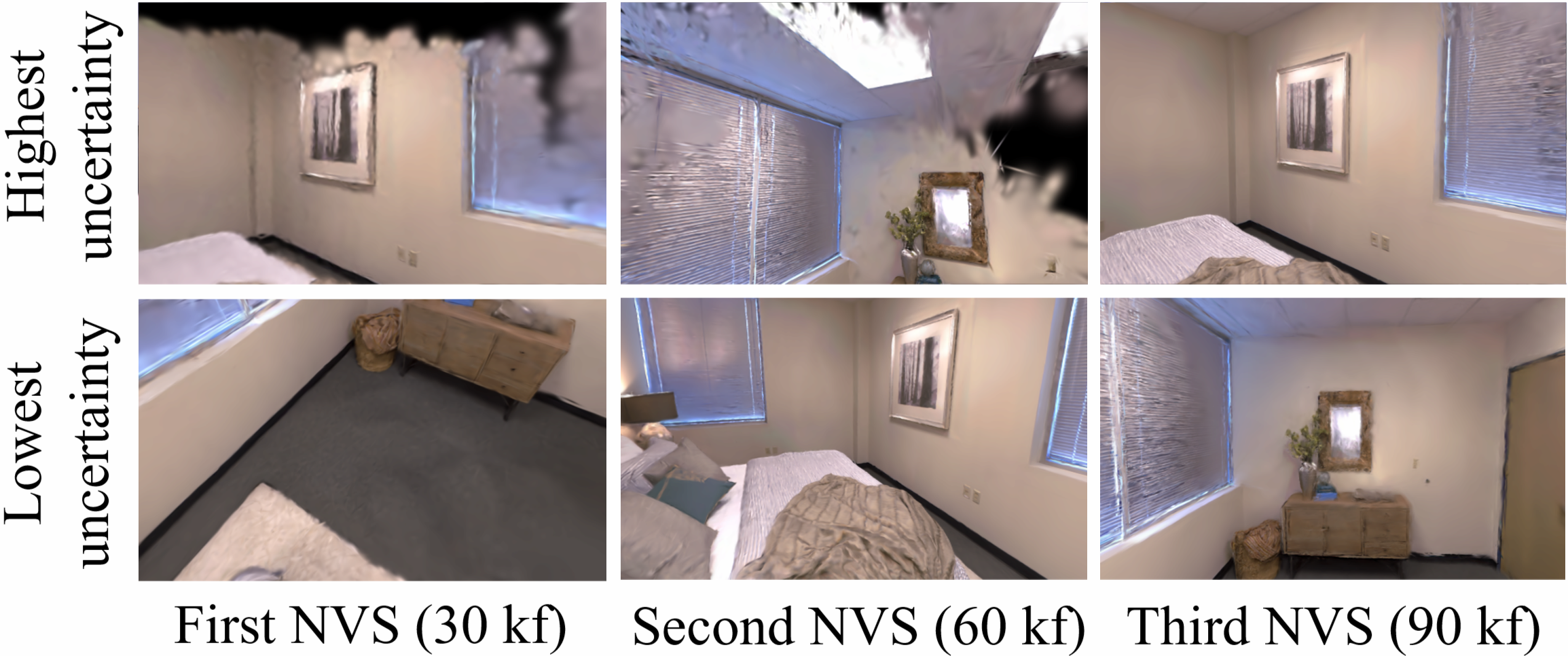}  
    \caption{Rendered images from views with the highest and lowest information gain in the candidate set at every 30th keyframe.}
    \label{fig3: high low uncertainty comparison}
\end{figure}

Using these two components, an uncertainty score $U_{n,i}$ is computed for each Gaussian $g_{n,i}$ as follows:  
\begin{equation}
    U_{n,i}=\alpha_1\lambda_{n,i}+\alpha_2A_{n,i}
    \tag{7}
\end{equation}
\noindent where $\alpha_1=0.7,$ $\alpha_2=0.3$ are weighted hyperparameters.

Finally, the information gain of the view $I_n$ is defined as the total uncertainty of all Gaussians visible from $I_n$.
\begin{equation}
    U_n = \sum_{i=1}^{\left|\left\{g_{n,i}\right\}\right|} \frac{1}{z_{n,i}^2} U_{n,i} \tag{8}
\end{equation}
\noindent where $z_{n,i}$ represents the depth of the Gaussian $g_{n,i}$ in the view $I_n$. This equation highlights that Gaussians closer to the view $I_n$ have a greater impact on the uncertainty score.

After calculating an information gain for each non-keyframe, we sort all candidate frames in descending order based on their scores. Since consecutive frames often have similar scores, we apply non-maximum suppression (NMS) \cite{redmon2016you} within a local window to the sorted list. 
This process avoids selecting frames within the local window of previously chosen frames, promoting the selection of frames from diverse viewpoints. The top-k frames with the highest information gain are then selected and combined with the 30 most recently tracked keyframes to form the training frame set.
The frame set is used for additional training, enabling the 3DGS model to enhance under-reconstructed areas and refine recently observed regions that require further training.

We construct a candidate set consisting of non-keyframes captured within the span of the 30 most recent keyframes, and apply NVS to this set.
We identify 20 high-information-gain frames selected from the current set and include them in the subsequent candidate set. This approach allows for an efficient evaluation of information gain by focusing on the latest non-keyframes while incorporating selected older non-keyframes. Fig.~3 depicts the non-keyframes with the highest and lowest gain within the set constructed at each iteration.


\noindent \textbf{Gaussian Training}
\noindent This optimization step is iteratively performed to refine Gaussian parameters. These Gaussians are trained using both keyframes and selected non-keyframes. The loss function for keyframes is defined as follows:
\begin{align*}
    L_{\mathrm{KF}} &= \lambda_{L1} L_1 + \lambda_{\mathrm{SSIM}} L_{\mathrm{SSIM}} \tag{9} \\ 
    &\quad + \lambda_{\mathrm{depth}} L_{\mathrm{depth}} + \lambda_{\mathrm{smooth}} L_{\mathrm{smooth}}
\end{align*}
where  $L_1$ represents the L1 loss between the rendered images (Eq. 2) and the original images. $L_{\mathrm{SSIM}}$ is the structural similarity index measure (SSIM) loss, which preserves image quality \protect\cite{1284395}. $L_{\mathrm{depth}}$ is calculated between the rendered depths (Eq. 3) and the MVS depths, helping to maintain the geometric structure. Lastly $L_{\mathrm{smooth}}$ is the smoothness loss, designed to minimize depth differences between neighboring pixels, which is expressed as:  
\vspace{-2pt}
\begin{align*}
L_{\mathrm{smooth}} &= \frac{1}{N_x} \sum_{i,j} \left| d(i,j) - d(i,j+1) \right| \tag{10} \\
&\quad + \frac{1}{N_y} \sum_{i,j} \left| d(i,j) - d(i+1,j) \right|
\end{align*}
This smooths abrupt depth changes between pixels, yielding natural transitions and fewer discontinuities.

Non-keyframes contain only image information and do not include depth information. Therefore, for the non-keyframes, the loss function simplifies image rendering losses:
\begin{equation}
L_{\mathrm{NKF}}=\lambda_{L1} L_1+\lambda_{\mathrm{SSIM}} L_{\mathrm{SSIM}}+\lambda_{\mathrm{smooth}} L_{\mathrm{smooth}} \tag{11}
\end{equation}
\noindent This approach ensures that both keyframes and non-keyframes contribute effectively to improving the 3DGS model, balancing geometric accuracy and image quality.

\begin{figure}[t]
    \centering
    \begin{subfigure}[b]{0.47\textwidth}
        \includegraphics[width=\linewidth]{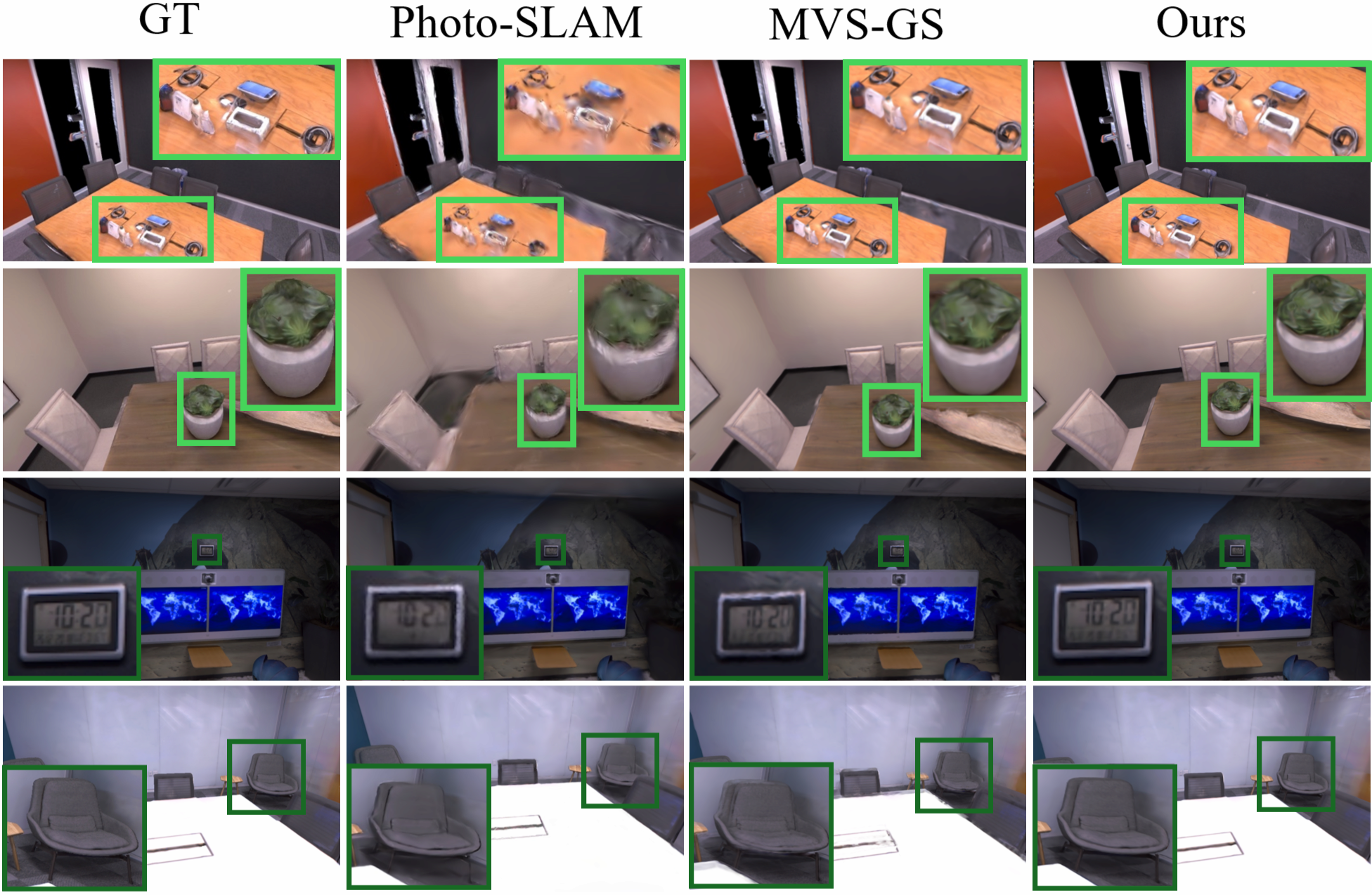}
        \caption{ Results on the Replica dataset }
        \label{fig:subfig_a}
        \vspace{5pt}
    \end{subfigure}
    \hfill
    \begin{subfigure}[b]{0.47\textwidth}
        \includegraphics[width=\linewidth]{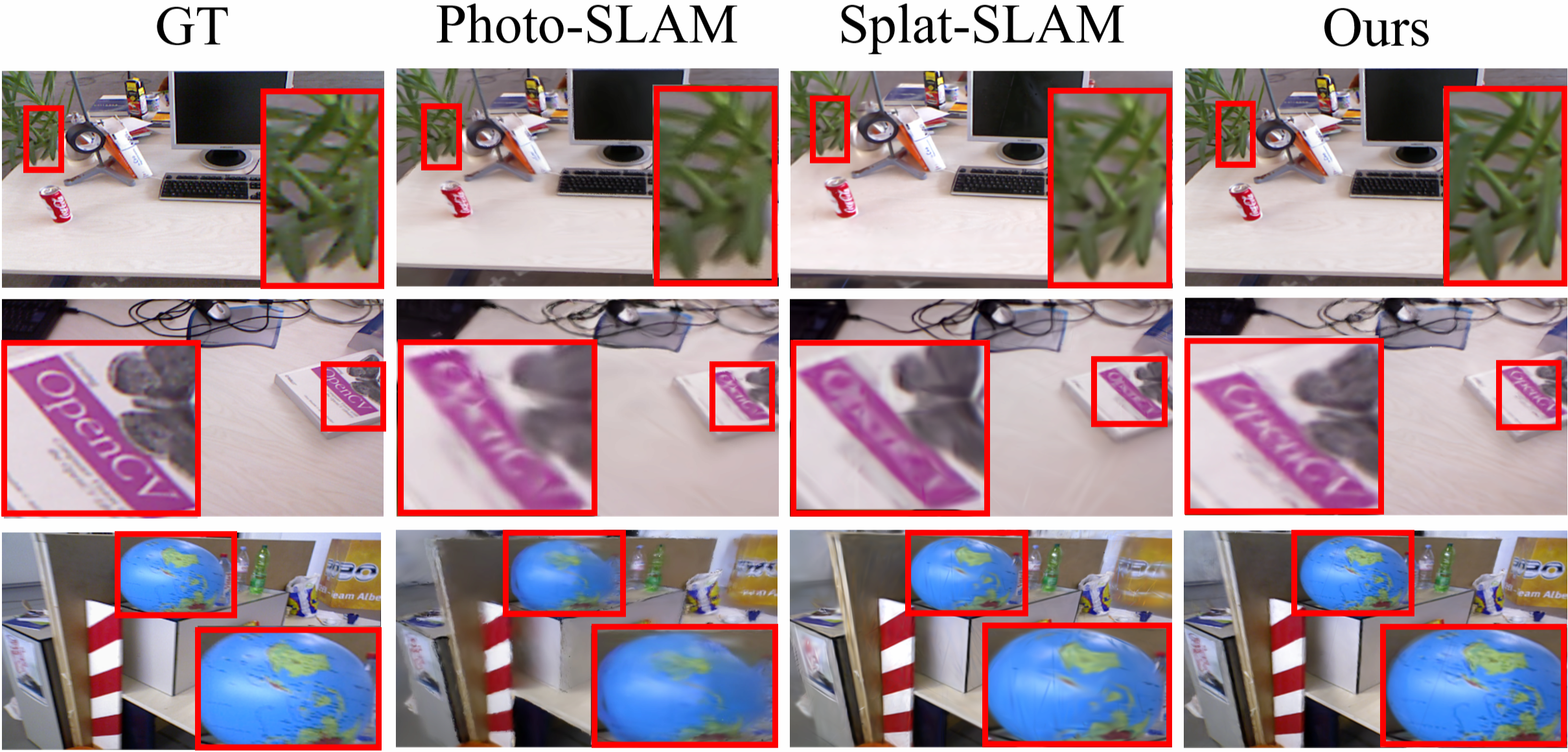}
        \caption{ Results on the TUM-RGBD dataset }
        \label{fig:subfig_b}
    \end{subfigure}
    \caption{Qualitative evaluation on (a) the Replica and (b) TUM-RGBD datasets. The boxes highlight areas where our method produces fine details with higher accuracy than the competing methods.}
    \label{fig:main_figure}
\end{figure}

\section{Experiment Results}

To assess the quality of our method, we compared to state-of-the-art methods on various datasets, including indoor and outdoor scenes. For indoor scenes, we conducted experiments using the Replica \protect\cite{straub2019replica} and TUM-RGBD \protect\cite{sturm2012benchmark} datasets, following the same experimental setups as other methods (\protect\cite{matsuki2024gaussian}, \protect\cite{sandstrom2024splat}, \protect\cite{zhang2024glorie}). For outdoor scenarios, we used the Aerial \protect\cite{song2021view} and Tanks\&Temples datasets \protect\cite{knapitsch2017tanks}).

\noindent\textbf{Implementation Details.} All experiments were carried out on a desktop equipped with an AMD Ryzen9 7900X 12-core processor and an NVIDIA GeForce RTX 4090 GPU. Training and evaluation were performed in PyTorch with CUDA to accelerate rasterization and gradient computations. While most hyperparameters follow the original 3DGS setting \cite{kerbl20233d}, we empirically set $\lambda_{L1}$, $\lambda_{SSIM}$, $\lambda_{depth}$ and $\lambda_{smooth}$ to 0.95, 0.2, 0.2, and 0.1, respectively, for the loss function of Gaussian training. We used 100 non-keyframes for NVS, and set the NMS gap between frames to 3 to exclude neighboring frames.

\noindent\textbf{Baseline Methods.}
We evaluated the performance of our method by comparing it with state-of-the-art monocular dense SLAM methods, including NeRF-based (Q-SLAM \protect\cite{peng2024q} and GLORIE-SLAM \protect\cite{zhang2024glorie}) and 3DGS-based (MonoGS \protect\cite{matsuki2024gaussian}, Splat-SLAM \protect\cite{sandstrom2024splat}, Photo-SLAM \protect\cite{huang2024photo}, and MVS-GS \protect\cite{lee2024mvs}) methods.

\begin{table}[t]
\centering
{\small 
\resizebox{0.48\textwidth}{!}{%
\begin{tabular}{l cccccccc c}
\toprule
\textbf{Method} & \textbf{Off0} & \textbf{Off1} & \textbf{Off2} & \textbf{Off3} & \textbf{Off4} & \textbf{Rm0} & \textbf{Rm1} & \textbf{Rm2} & \textbf{Avg.} \\ \midrule
\multicolumn{10}{c}{\textbf{PSNR $\uparrow$}} \\ \midrule
Q-SLAM      & 36.31 & 37.22 & 30.68 & 30.21 & 31.96 & 29.58 & 32.74 & 31.25 & 32.49 \\
MonoGS      & 32.00 & 31.21 & 23.26 & 25.77 & 23.85 & 23.53 & 25.00 & 22.42 & 25.88 \\
Splat-SLAM  & 40.81 & 40.64 & \underline{35.19} & \underline{35.03} & \underline{37.40} & \underline{32.25} & \underline{34.31} & \underline{35.95} & \underline{36.45} \\
Photo-SLAM  & 36.99 & 37.52 & 31.79 & 31.62 & 34.17 & 29.77 & 31.30 & 33.18 & 33.29 \\
MGS-SLAM    & 35.51 & 34.25 & 30.83 & 31.86 & 34.38 & 29.91 & 31.06 & 31.49 & 32.41 \\
MVS-GS      & \underline{41.02} & \underline{42.04} & 34.00 & 34.65 & 33.33 & 32.20 & 31.54 & 35.84 & 35.58 \\
Ours        & \textbf{43.93} & \textbf{43.98} & \textbf{37.98} & \textbf{36.31} & \textbf{39.59} & \textbf{34.88} & \textbf{37.99} & \textbf{39.60} & \textbf{39.28} \\ \midrule
\multicolumn{10}{c}{\textbf{SSIM $\uparrow$}} \\ \midrule
Q-SLAM      & 0.94 & 0.94 & 0.90 & 0.88 & 0.89 & 0.83 & 0.91 & 0.87 & 0.89 \\
MonoGS      & 0.90 & 0.88 & 0.82 & 0.84 & 0.86 & 0.75 & 0.79 & 0.81 & 0.83 \\
Splat-SLAM  & 0.97 & \textbf{0.99} & \underline{0.97} & \textbf{0.97} & \underline{0.97} & \textbf{0.96} & \textbf{0.97} & \underline{0.96} & \underline{0.97} \\
Photo-SLAM  & 0.96 & 0.95 & 0.93 & 0.92 & 0.94 & 0.87 & 0.91 & 0.93 & 0.93 \\
MGS-SLAM    & 0.94 & 0.93 & 0.90 & 0.92 & 0.95 & 0.89 & 0.90 & 0.91 & 0.92 \\
MVS-GS      & \underline{0.98} & \underline{0.98} & 0.95 & \underline{0.96} & 0.95 & \underline{0.95} & \underline{0.92} & \underline{0.96} & 0.96 \\
Ours        & \textbf{0.99} & \textbf{0.99} & \textbf{0.98} & \textbf{0.97} & \textbf{0.98} & \textbf{0.96} & \textbf{0.97} & \textbf{0.98} & \textbf{0.98} \\ \midrule
\multicolumn{10}{c}{\textbf{LPIPS $\downarrow$}} \\ \midrule
Q-SLAM      & 0.13 & 0.15 & 0.20 & 0.19 & 0.18 & 0.18 & 0.16 & 0.15 & 0.17 \\
MonoGS      & 0.23 & 0.22 & 0.30 & 0.24 & 0.34 & 0.33 & 0.35 & 0.39 & 0.30 \\
Splat-SLAM  & \underline{0.05} & 0.07 & \underline{0.06} & \underline{0.04} & 0.10 & 0.09 & \underline{0.06} & \underline{0.05} & \underline{0.06} \\
Photo-SLAM  & 0.06 & 0.06 & 0.09 & 0.09 & \underline{0.07} & 0.10 & 0.08 & 0.07 & 0.08 \\
MGS-SLAM    & 0.07 & 0.11 & 0.12 & 0.07 & 0.08 & \underline{0.08} & 0.09 & 0.09 & 0.09 \\
MVS-GS      & \underline{0.05} & \underline{0.05} & 0.09 & 0.07 & 0.10 & 0.10 & 0.13 & 0.07 & 0.08 \\
Ours        & \textbf{0.02} & \textbf{0.02} & \textbf{0.03} & \textbf{0.03} & \textbf{0.03} & \textbf{0.04} & \textbf{0.04} & \textbf{0.03} & \textbf{0.03} \\ \bottomrule
\end{tabular}%
}
\caption{Quantitative rendering performance on the Replica dataset.}
\label{tab:replica_performance}
}
\end{table}

\begin{table}[t]
\centering
{\tiny 
\resizebox{0.46\textwidth}{!}{%
\begin{tabular}{c c c c c c}
\toprule
\textbf{Method} & \textbf{Metrics} & \textbf{f1/desk} & \textbf{f2/xyz} & \textbf{f3/off} & \textbf{Avg.} \\ \midrule
MonoGS       & \multirow{6}{*}{PSNR $\uparrow$} & 19.67 & 16.17 & 20.63 & 18.82 \\
Photo-SLAM   &                                  & 20.97 & 21.07 & 19.59 & 20.54 \\
GLORIE-SLAM  &                                  & 20.26 & 25.62 & 21.21 & 22.36 \\
Splat-SLAM   &                                  & \underline{25.61} & \textbf{29.53} & \underline{26.05} & \underline{27.06} \\
MVS-GS       &                                  & 20.67 & 24.53 & 22.37 & 22.52 \\
Ours         &                                  & \textbf{26.34} & \underline{28.72} & \textbf{28.10} & \textbf{27.72} \\ \midrule
MonoGS       & \multirow{6}{*}{SSIM $\uparrow$} & 0.73 & 0.72 & 0.77 & 0.74 \\
Photo-SLAM   &                                  & 0.74 & 0.73 & 0.69 & 0.72 \\
GLORIE-SLAM  &                                  & \underline{0.87} & \textbf{0.96} & \underline{0.84} & \underline{0.89} \\
Splat-SLAM   &                                  & 0.84 & 0.90 & \underline{0.84} & 0.86 \\
MVS-GS       &                                  & 0.77 & 0.86 & 0.80 & 0.81 \\
Ours         &                                  & \textbf{0.89} & \underline{0.91} & \textbf{0.91} & \textbf{0.90} \\ \midrule
MonoGS       & \multirow{6}{*}{LPIPS $\downarrow$} 

& 0.33 & 0.31 & 0.34 & 0.33 \\
Photo-SLAM   &                                      & 0.23 & 0.17 & 0.24 & 0.21 \\
GLORIE-SLAM  &                                      & 0.31 & 0.09 & 0.32 & 0.24 \\
Splat-SLAM   &                                      & \underline{0.18} & \underline{0.08} & \underline{0.20} & \underline{0.15} \\
MVS-GS       &                                      & 0.25 & 0.15 & 0.24 & 0.21 \\
Ours         &                                      & \textbf{0.12} & \textbf{0.07} & \textbf{0.10} & \textbf{0.10} \\ \bottomrule
\end{tabular}%
}
\caption{Quantitative rendering performance on the TUM-RGBD.}
\label{tab:tum_rgbd}
}
\end{table}

\noindent\textbf{Evaluation Metrics.}
We assessed the modeling quality using rendered images and depth maps. For image rendering quality, we report PSNR, SSIM, and LPIPS by comparing the rendered images with the original images.

\subsection{Evaluation in Indoor Scenes}

Table 1 shows the rendering performance results on the Replica dataset, highlighting that our method significantly outperforms other methods. Notably, it achieves superior results in both PSNR and SSIM, with a substantial improvement compared to the second-best method, Splat-SLAM.

Fig.~4a shows a qualitative comparison between our method and the other 3DGS-based methods. Our method effectively captures intricate scene details, resulting in high-quality renderings with fewer artifacts. Our texture details are significantly better than those produced by Photo-SLAM. Notably, MVS-GS fails to handle loop closing, causing the chair to appear doubled. In contrast, our method accurately reconstructs a single, complete representation of the chair. 

Table 2 presents the evaluation results on real-world scenes from the TUM-RGBD dataset. Despite the inherent challenges of real-world datasets, including noisy images and complex scene variations, our approach maintains stable and reliable rendering quality. The qualitative results are shown in Fig.~4b. This demonstrates the robustness of our method in handling real-world data, ensuring consistent performance across diverse indoor scenes.

\subsection{Evaluation in Outdoor Scenes}


In this section, we validated the generalization capability of our method by evaluating its performance on outdoor scenes. Depth prediction-based methods, including Splat-SLAM and GLORIE-SLAM, failed to generate 3DGS models in these environments. Fig.~5a presents a performance evaluation of aerial scenes, demonstrating that our method outperforms both Photo-SLAM and MVS-GS, particularly in capturing small structural details. Leveraging the strengths of MVS in accurately reconstructing large-scale scenes, our method significantly outperformed Photo-SLAM in aerial scenes.


On the Tanks and Temples dataset, Splat-SLAM, MonoGS, and even Photo-SLAM failed to generate 3DGS models due to the lack of sufficient continuous motion in the image frames. Fig.~5b presents the rendering results for this dataset. While MVS-GS successfully produced a 3DGS model, its quality was compromised by limited training views and the absence of loop closure. In contrast, our method delivered high-quality rendering results that closely resembled real photographs. These findings highlight the effectiveness of the proposed NVS method and consistent geometric alignment, even in challenging outdoor scenes.

\begin{figure}[t]
    \centering
    \begin{subfigure}[b]{0.48\textwidth}
        \includegraphics[width=\textwidth]{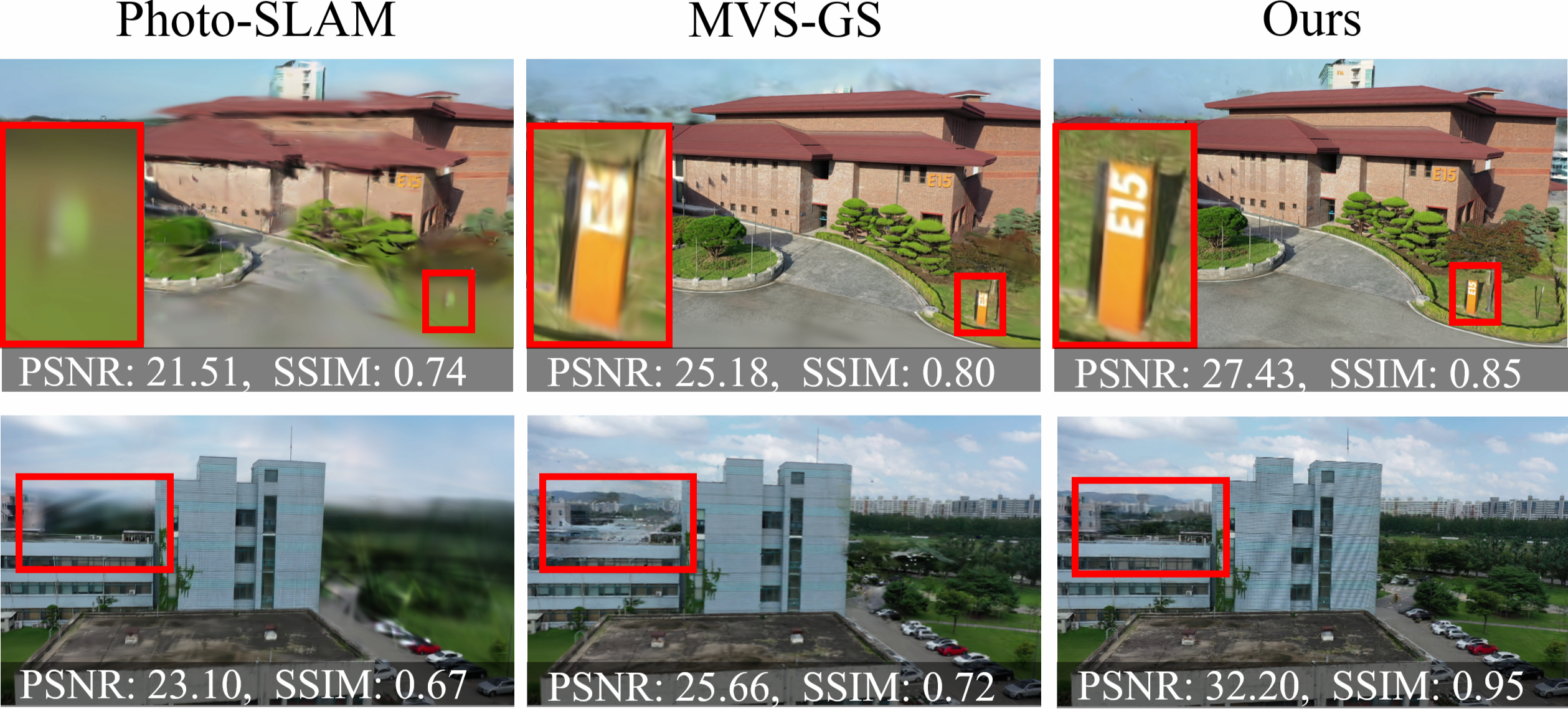}
        \caption{ Results on the Aerial dataset }
        \label{fig:qualitative comparison on Aerial Dataset}
        \vspace{5pt}
    \end{subfigure}
    \hfill
    \begin{subfigure}[b]{0.48\textwidth}
        \includegraphics[width=\textwidth]{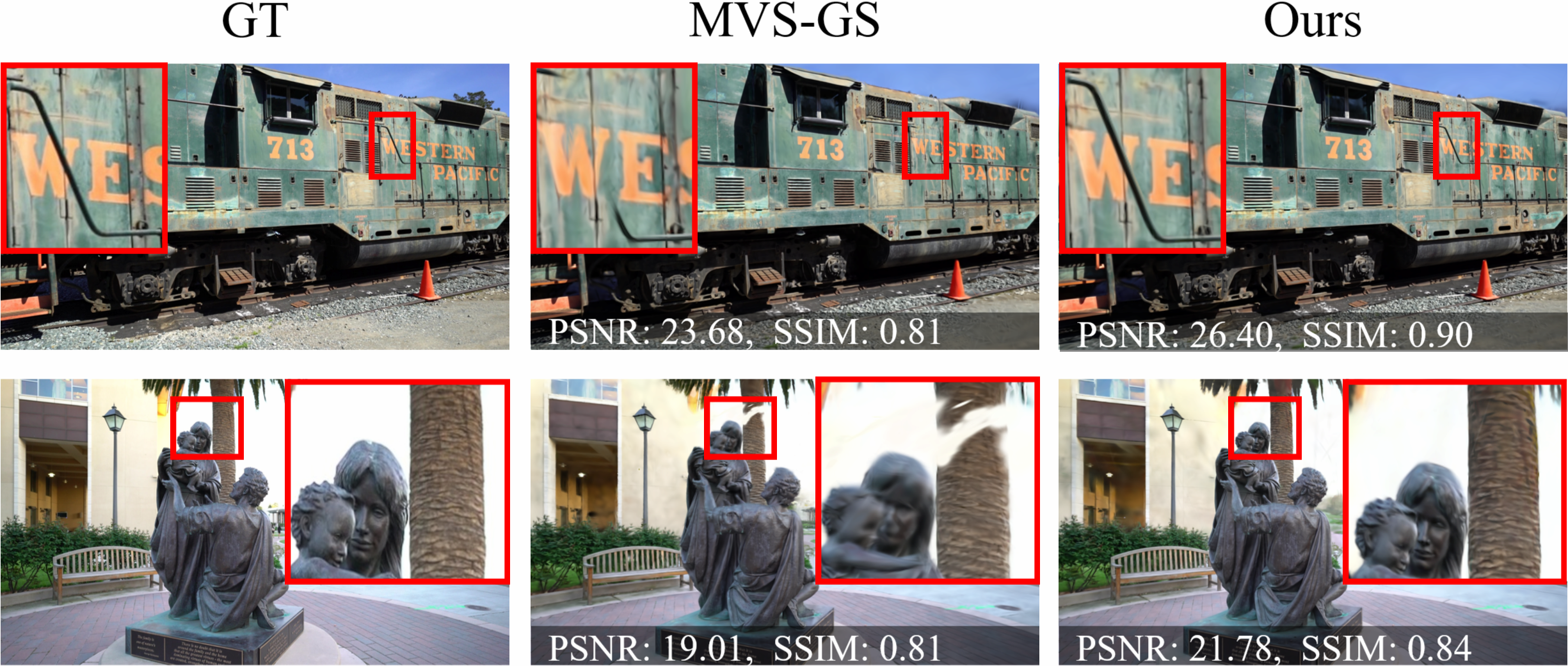}
        \caption{ Results on the Tanks and Temples dataset }
        \label{fig:qualitative result on Tanks and Temples dataset}
    \end{subfigure}
    \caption{Qualitative evaluation on (a) the aerial scenes and (b) Tanks and Temples datasets. The red boxes highlight areas where our method achieves better accuracy in capturing fine details.}
    \label{fig:main_figure}
\end{figure}



\subsection{Ablation Study}
To evaluate the impact of each key component in our method, we performed an ablation study on the ``Office0'' scene from the Replica dataset. Each component was progressively assessed as follows:
\begin{itemize}[leftmargin=0.3cm]
    \item \textbf{Method-A}: The baseline method, MVS-GS.
    \item \textbf{Method-B}: Incorporates the global GBA module.
    \item \textbf{Method-C}: Builds on Method-B by applying depth initialization using the disparity map from DROID-SLAM.
    \item \textbf{Method-D}: Extends Method-C by introducing a smoothness loss for training Gaussians.
    \item \textbf{Method-E}: Represents our full model, including the NVS.
\end{itemize}

We evaluated both image and depth rendering performance using metrics such as PSNR, SSIM, and LPIPS for images, and L1 error for depths.
We also reported the number of Gaussians, map size, and computational speed (measured in frames per second, FPS) for each method.
Table 3 summarizes the results for the variant methods.

\begin{table}[t]
\centering
\resizebox{0.48\textwidth}{!}{ 
\begin{tabular}{cccccccc}
\toprule
\makecell{\textbf{Method}} 
& \makecell{\textbf{PSNR $\uparrow$}} 
& \makecell{\textbf{SSIM $\uparrow$}} 
& \makecell{\textbf{LPIPS $\downarrow$}} 
& \makecell{\textbf{Depth} \textbf{L1} $\downarrow$}
& \makecell{\textbf{FPS $\uparrow$}}
& \makecell{\textbf{\#Gaussians} \textbf{(K) $\downarrow$}} 
& \makecell{\textbf{Map Size} \\ \textbf{[GiB] $\downarrow$}} \\
\toprule
A & 40.92 & 0.979 & 0.023 & 0.042 & 19.39 & 1365 & 0.305 \\
B & 42.37 & 0.984 & 0.019 & 0.046 & 15.66 & 1253 & 0.280 \\
C & 42.71 & 0.985 & 0.018 & 0.044 & 15.28 & 1357 & 0.303 \\
D & 42.73 & 0.985 & 0.019 & 0.038 & 15.11 & 1377 & 0.308 \\
E & 43.93 & 0.988 & 0.016 & 0.034 & 9.18 & 1078 & 0.241 \\
\midrule
\end{tabular}
}
\caption{Comparison results of our method with different model variants on the Replica ``office 0'' scene.}
\label{Table_3}
\end{table}

\begin{figure}[t]  
\centering
\includegraphics[width=0.475\textwidth]{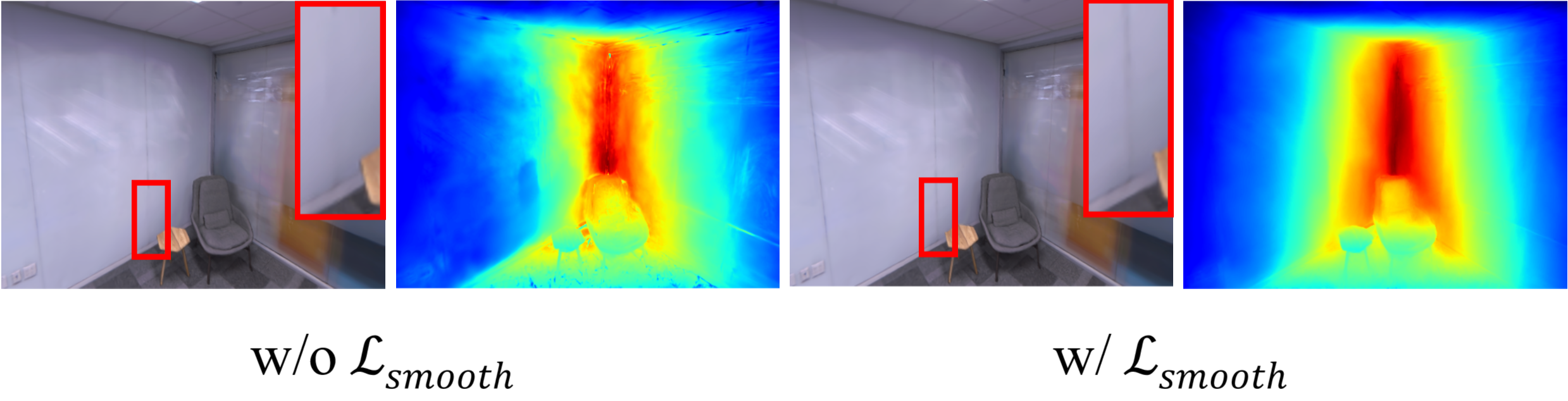}
\caption{The effect of depth smooth regularization loss on the rendered image and depth map.}
\label{fig:depth smooth loss ablation result}  
\end{figure}


\begin{table}[t]
\centering
\resizebox{0.48\textwidth}{!}{ 
\begin{tabular}{cccccccc}
\toprule
Metrics
& \makecell{$\alpha_1$=0.0 \\ $\alpha_2$=1.0} 
& \makecell{$\alpha_1$=0.1 \\ $\alpha_2$=0.9} 
& \makecell{$\alpha_1$=0.3 \\ $\alpha_2$=0.7} 
& \makecell{$\alpha_1$=0.5 \\ $\alpha_2$=0.5} 
& \makecell{$\alpha_1$=0.7 \\ $\alpha_2$=0.3} 
& \makecell{$\alpha_1$=0.9 \\ $\alpha_2$=0.1} 
& \makecell{$\alpha_1$=1.0 \\ $\alpha_2$=0.0} \\
\midrule
PSNR $\uparrow$ & 43.802 & 43.820 & 43.859 & 43.877 & \textbf{43.932} & 43.841 & 43.812 \\
SSIM $\uparrow$ & 0.986  & 0.986  & 0.987  & 0.987  & \textbf{0.988}  & 0.987  & 0.985  \\
LPIPS $\downarrow$ & 0.018  & 0.017  & 0.017  & 0.016  & \textbf{0.016}  & 0.018  & 0.019  \\
\bottomrule
\end{tabular}
}
\caption{Performance variations with respect to changes in the weights $\alpha_1$ and $\alpha_2$ of the uncertainty on the Replica ``office 0''.}
\label{tab:system_performance}
\end{table}

Compared to Method-A, both Method-B and Method-C showed progressive improvements in rendering performance. This indicates that the application of the GBA and the depth initialization approach enhances the completeness of 3DGS modeling. Furthermore, Method-D incorporated the smooth depth loss, which showed no noticeable improvement in rendering quality but led to substantial enhancements in reconstruction performance. Fig.~6 shows the comparison results between Method-C and Method-D, where the improvements in reconstruction performance are visually apparent. Finally, when the NVS technique was applied, Method-E achieved superior performance in both rendering quality and depth reconstruction. Furthermore, in terms of model compactness, Method-E significantly reduces the number of Gaussians compared to Method-D, resulting in approximately a 60 MB reduction in map size. The NVS optimization helps eliminate or merge redundant keyframe Gaussians, thereby minimizing unnecessary overhead.

We further analyzed the impact of the uncertainty score parameters in Eq. (7) on the effectiveness of view selection. Table 4 shows the effect of the $\alpha_1$ and $\alpha_2$. The largest eigenvalue primarily captures uncertainty in high-frequency regions, while the positional gradient focuses on low-frequency areas. Thus, $\alpha_1$ and $\alpha_2$ serve to balance these two complementary measures. If either value is too large, uncertainty estimation becomes biased and less effective. Therefore, we empirically determined suitable $\alpha_1$ and $\alpha_2$ values for balanced and accurate estimation.

\section{Conclusion}
We present a novel online mapping approach for high-quality 3DGS modeling that introduces uncertainty estimation directly within Gaussians. Our method utilizes an uncertainty-aware view selection strategy to extract the most informative views, improving the completeness of 3DGS models. Additionally, we develop a framework that integrates an online MVS technique, maintaining consistency in 3D information throughout the modeling process. Extensive experiments show that the proposed approach surpasses state-of-the-art dense SLAM methods, demonstrating outstanding performance, especially in complex outdoor environments. To the best of our knowledge, this is the first work to introduce non-keyframe selection in a 3DGS-based SLAM framework. The proposed formulation is general and applicable to various SLAM systems.

\section*{Acknowledgments}
This research was supported by the MSIT(Ministry of Science and ICT), Korea, under the ITRC(Information Technology Research Center) support program(IITP-2025-RS-2020-II201789), and the Artificial Intelligence Convergence Innovation Human Resources Development(IITP-2025-RS-2023-00254592) supervised by the IITP(Institute for Information \& Communications Technology Planning \& Evaluation).


\clearpage

\setcounter{section}{0}
\renewcommand{\thesection}{\Alph{section}}
\setcounter{table}{0}
\renewcommand{\thetable}{S\arabic{table}}
\setcounter{figure}{0}
\renewcommand{\thefigure}{S\arabic{figure}}

\textbf{\huge Appendix}
\vspace{-3pt}

\setcounter{subsection}{1}

\section{Detailed Method}
\subsection{Camera Tracking}
We utilized DROID-SLAM \cite{teed2021droid} to track camera poses of input image stream. The core component of DROID-SLAM is an update operator that iteratively refines camera pose and disparity map of each keyframe. Specifically, for each keyframe pair $\left(I_k,I_l\right)$, where $I_k$ is the incoming keyframe to be tracked and $I_l$ is a neighboring keyframe selected from the existing frame graph $G$, we first extract feature maps and construct a correlation cost volume $C_{k,l}$.  This volume is then used as input for the update operator. During each iteration, the update operator employs ConvGRU recurrent layers and dense bundle adjustment (DBA) layers to refine the pose and disparity estimates. After several iterations, the outputs for keyframe $I_k$ converge to a stable solution, allowing the keyframe to be added to the frame graph.

\subsection{MVS Depth Estimation}
When keyframe $I_k$ was tracked, we subsequently perform multi-view depth estimation to obtain an accurate depth map for this keyframe $I_k$.  To do so, we utilize a learning-based MVS method, MVSFormer \cite{cao2022mvsformer}. This is a coarse-to-fine architecture, which first estimates a depth map at a low image resolution and then gradually refines the depth in finer resolution.
In our implementation, we find that the stability of initial depth map is crucial in the above process. Therefore, we leverage the disparity map estimated by the DROID-SLAM to initialize a depth map for MVSFormer. This also improves the model efficiency at the same time. However, it is observed that the disparity map usually exhibits noisy artifacts at boundary regions. To address this, we replace pixel values at the boundaries to average values of their neighboring pixels. This approach enhances the quality of initial depth map and provides a robust foundation for generating more accurate depth maps in subsequent stages.

\subsection{Global Dense Bundle Adjustment}
Our global DBA module is executed every 30 keyframes to optimize camera poses and disparities across the entire frame graph $G$. To ensure numerical stability, the disparities and poses are normalized before each optimization step. Specifically, the average disparity $\bar{d}$ is calculated across all keyframes. Then, each disparity of each keyframe is normalized by $d_{norm}=d/\bar{d}$. For camera pose normalization, the translation vector is computed as $t_{norm}=\bar{d}t$.

Assume that a camera pose $T_{k}$ is updated to a new pose $T_{k}^{'}$ after GBA. To facilitate deformation, a relative transformation is computed as $T_{k}^{rel}=T_{k}^{'}T_{k}^{-1}$. After that, the mean $\mu_{i}$ and covariance $\sum_{i}$ of each Gaussian $g_{i}$ is transformed accordingly as follows:
\begin{equation}
\mu_i^\prime=T_k^{rel}\mu_i, \; R_i^\prime=T_k^{rel}R_i
\tag{4}
\end{equation}
\begin{equation}
\Sigma_\mathrm{i}^\prime=R_k^{rel}\Sigma_\mathrm{i}\left(R_k^{rel}\right)^\top
\tag{5}
\end{equation}
\noindent where $R_{k}^{rel}$ is the rotation component of camera pose $T_{k}^{rel}$.
Note that the means $\mu_{i}$, $\mu_{i}^{'}$ are written in homogeneous coordinate.

\subsection{Pruning and Densification}
During the 3DGS mapping, we adopt the pruning and densification strategies outlined in \cite{hamdi2024ges}. For pruning, Gaussians with an opacity below 0.005 are globally removed every 150 mapping iterations. Additionally, Gaussians with excessively large scales in both 2D and 3D projections are also eliminated. Densification is performed every 150 mapping iterations based on accumulated gradients. The gradient threshold for densification is set to 0.0002. Furthermore, during the Novel View Training process, densification is selectively applied to Gaussians with high uncertainty. All Gaussians are periodically reset following the method described in \cite{hamdi2024ges}, with opacity being reinitialized every 3000 mapping iterations and shape parameters being reinitialized every 100 iterations.

\subsection{Note on Rendering Evaluation}
To ensure a fair comparison across all methods, we conducted at least five experiments and averaged the results for evaluation. Additionally, for methods that do not render depth maps as described in \cite{kerbl20233d}, we applied a consistent implementation to ensure uniform evaluation.

\section{More Experiments}
\subsection{Implementation Details}

In all frames, we apply a voxel downsampling factor of $0.005$. For the Replica dataset \cite{straub2019replica}, the final refinement process during Gaussian training uses $\beta$ = 2000 iterations. For the TUM-RGBD \cite{sturm2012benchmark} and ScanNet\cite{dai2017scannet} datasets, $\beta$ is adjusted to 26000, following the configurations used in other methods (\cite{matsuki2024gaussian}, \cite{sandstrom2024splat}, \cite{zhang2024glorie}).
We adopted the hyperparameter settings specified in \cite{hamdi2024ges} for Gaussian mapping and \cite{teed2021droid} for tracking.

\subsection{Quantitative Performance on the ScanNet}
To further assess rendering performance, we conducted additional evaluations using the ScanNet dataset \cite{dai2017scannet}. The results, summarized in Table S1, highlight the comparative performance of our method.

\begin{table}[ht]
\centering
{
\resizebox{0.48\textwidth}{!}{%
\begin{tabular}{l l ccccccc}
\toprule
& \textbf{Method} & \textbf{0000} & \textbf{0059} & \textbf{0106} & \textbf{0169} & \textbf{0181} & \textbf{0207} & \textbf{Avg.} \\ \midrule
\multirow{5}{*}{\rotatebox{90}{PSNR $\uparrow$}} & MonoGS       &  16.91 & 19.15 & 18.57 & 20.21 & 19.51 & 18.37 & 18.79 \\
&GLORIE-SLAM  &   23.42 & 20.66 & 20.41 & 25.23 & 21.28 & 23.68 & 22.45 \\
&Splat-SLAM   &   \underline{28.68} & \underline{27.69} & \underline{27.70} & \underline{31.14} & \textbf{31.15} & \underline{30.49} & \underline{29.48} \\
&MVS-GS       &   23.91 & 22.60 & 21.65 & 25.56 & 22.18 & 24.22 & 23.35 \\
&Ours         &   \textbf{29.06} & \textbf{29.22} & \textbf{27.88} & \textbf{32.42} & \underline{29.48} & \textbf{30.68} & \textbf{29.79} \\ \midrule
\multirow{5}{*}{\rotatebox{90}{SSIM $\uparrow$}}  & MonoGS       &  0.62 & 0.69 & 0.74 & 0.74 & 0.75 & 0.70 & 0.71 \\
&GLORIE-SLAM  &   \underline{0.87} & 0.87 & 0.83 & 0.84 & \textbf{0.91} & 0.76 & \underline{0.85} \\
&Splat-SLAM   &   0.83 & 0.87 & \underline{0.86} & \underline{0.87} & 0.84 & \underline{0.84} & \underline{0.85} \\
&MVS-GS       &   0.82 & \underline{0.88} & 0.84 & 0.85 & 0.80 & 0.77 & 0.83 \\
&Ours         &   \textbf{0.88} & \textbf{0.91} & \textbf{0.90} & \textbf{0.91} & \underline{0.90} & \textbf{0.88} & \textbf{0.90} \\ \midrule
\multirow{5}{*}{\rotatebox{90}{LPIPS $\downarrow$}}  & MonoGS       &  0.70 & 0.51 & 0.55 & 0.54 & 0.63 & 0.58 & 0.59 \\
&GLORIE-SLAM  &   0.26 & 0.31 & 0.31 & 0.21 & 0.44 & 0.29 & 0.30 \\
&Splat-SLAM   &   \underline{0.19} & \textbf{0.15} & \textbf{0.18} & \textbf{0.15} & \textbf{0.23} & \textbf{0.19} & \textbf{0.18} \\
&MVS-GS       &   0.24 & \underline{0.22} & 0.27 & 0.21 & 0.37 & 0.27 & 0.26 \\
&Ours         &   \textbf{0.14} & \textbf{0.15} & \underline{0.22} & \underline{0.16} & \underline{0.27} & \underline{0.21} & \underline{0.19} \\
\bottomrule
\end{tabular}%
}
\caption{Quantitative rendering performance on the ScanNet.}
}
\end{table}

\clearpage 
\begin{table*}[t]
\renewcommand{\arraystretch}{1.5} 
\setlength{\tabcolsep}{10pt} 
\centering
\small{
\resizebox{1\textwidth}{!}{
\begin{tabular}{cccccccccccc}
\toprule
\textbf{} & \multicolumn{2}{c}{\textbf{Metrics}} & \textbf{O-0} & \textbf{O-1} & \textbf{O-2} & \textbf{O-3} & \textbf{O-4} & \textbf{R-0} & \textbf{R-1} & \textbf{R-2} & \textbf{Avg.} \\ \midrule
 & Keyframes & Depth L1 $\downarrow$ & 0.034 & 0.055 & 0.049 & 0.053 & 0.046 & 0.060 & 0.048 & 0.054 & \textbf{0.050} \\ \midrule
\multirow{3}{*}{Ours} 
 & \multirow{3}{*}{Every 5 Frames} & PSNR $\uparrow$ & 42.47 & 42.39 & 36.49 & 35.72 & 37.84 & 33.89 & 36.32 & 37.67 & \textbf{37.85} \\
 &                        & SSIM $\uparrow$ & 0.98  & 0.98  & 0.97  & 0.97  & 0.97  & 0.95  & 0.97  & 0.97  & \textbf{0.97} \\
 &                        & LPIPS $\downarrow$ & 0.01 & 0.02 & 0.03 & 0.03 & 0.03 & 0.04 & 0.04 & 0.03 & \textbf{0.03} \\ \hdashline 
\multirow{3}{*}{GLORIE-SLAM} 
 & \multirow{3}{*}{Every 5 Frames} & PSNR $\uparrow$ & 27.15 & 28.85 & 28.56 & 33.95 & 36.27 & 26.78 & 26.95 & 28.11 & \textbf{29.58} \\
 &                        & SSIM $\uparrow$ & 0.95  & 0.96  & 0.95  & 0.97  & 0.99  & 0.96  & 0.96  & 0.95  & \textbf{0.96} \\
 &                        & LPIPS $\downarrow$ & 0.15 & 0.12 & 0.16 & 0.11 & 0.08 & 0.17 & 0.13 & 0.17 & \textbf{0.14} \\ \midrule
\multicolumn{3}{c}{Avg. FPS $\uparrow$}         & 8.70  & 11.94 & 8.66  & 7.96  & 8.67  & 10.77 & 9.14  & 7.61  & \textbf{9.18} \\
\multicolumn{3}{c}{GPU Usage [GiB] $\downarrow$} & 16.52 & 16.78 & 16.12 & 16.89 & 18.74 & 16.34 & 17.03 & 18.25 & \textbf{17.24} \\
\multicolumn{3}{c}{Number of Gaussians (1000x)} & 1035  & 921   & 1062  & 1624  & 970   & 1472  & 1067  & 1290  & \textbf{1183} \\
\bottomrule
\end{tabular}
}
\caption{Full Evaluation on Replica  \protect\cite{straub2019replica} For keyframes, performance is measured only on the frames used for mapping, following the same approach as existing methods. Additionally, rendering results are provided every 5 frames, regardless of whether they are keyframes or not. These results are labeled as "Every 5 Frames" and evaluated under the same conditions as other method \protect\cite{zhang2024glorie}. We report not only the depth evaluation results for keyframes and every 5 frames but also the maximum GPU usage and FPS.}
}
\end{table*}
\begin{table*}[t]
{\tiny
\centering
\renewcommand{\arraystretch}{1.5}
\setlength{\tabcolsep}{20pt}
\resizebox{1\textwidth}{!}{
\small 
\begin{tabular}{ccccccc}
\toprule
 & \multicolumn{2}{c}{\textbf{Metrics}} & \textbf{f1\_desk} & \textbf{f2\_xyz} & \textbf{f3\_office} & \textbf{Avg.} \\ \midrule
 & Keyframes & Depth L1 $\downarrow$ & 0.121 & 0.313 & 0.241 & \textbf{0.225} \\ \midrule
\multirow{3}{*}{Ours} 
& \multirow{3}{*}{Every 5 Frames} & PSNR $\uparrow$ & 23.59 & 22.79 & 25.44 & \textbf{23.94} \\
& & SSIM $\uparrow$ & 0.81 & 0.77 & 0.86 & \textbf{0.81} \\
& & LPIPS $\downarrow$ & 0.17 & 0.15 & 0.13 & \textbf{0.15} \\ \hdashline 
\multirow{3}{*}{GLORIE-SLAM} 
& \multirow{3}{*}{Every 5 Frames} & PSNR $\uparrow$ & 17.16 & 21.72 & 17.25 & \textbf{17.65} \\
& & SSIM $\uparrow$ & 0.79 & 0.92 & 0.73 & \textbf{0.77} \\
& & LPIPS $\downarrow$ & 0.40 & 0.15 & 0.44 & \textbf{0.37} \\ \midrule
\multicolumn{3}{c}{Avg. FPS $\uparrow$} & 1.41 & 7.77 & 4.76 & \textbf{4.65} \\
\multicolumn{3}{c}{GPU Usage [GiB] $\downarrow$} & 11.31 & 12.81 & 13.25 & \textbf{12.46} \\
\multicolumn{3}{c}{Number of Gaussians (1000x)} & 927 & 565 & 633 & \textbf{708} \\
\bottomrule
\end{tabular} 
}
\caption{Full Evaluation on TUM-RGBD \protect\cite{sturm2012benchmark}. The setups and evaluated metrics are similar to Table S1.}
}
\end{table*}
\begin{table*}[t]
\centering
\renewcommand{\arraystretch}{1.5} 
\resizebox{1\textwidth}{!}{
\begin{tabular}{cccccccccc}
\toprule
 & \multicolumn{2}{c}{\textbf{Metrics}} & \textbf{scene\_0000} & \textbf{scene\_0059} & \textbf{scene\_0106} & \textbf{scene\_0169} & \textbf{scene\_0181} & \textbf{scene\_0207} & \textbf{Avg.} \\ \midrule
 & Keyframes & Depth L1 $\downarrow$ & 0.169 & 0.278 & 0.439 & 0.216 & 0.208 & 0.183 & \textbf{0.248} \\ \midrule
\multirow{3}{*}{Ours} 
& \multirow{3}{*}{Every 5 Frames} & PSNR $\uparrow$ & 26.23 & 26.23 & 21.84 & 29.90 & 26.92 & 28.76 & \textbf{26.64} \\
& & SSIM $\uparrow$ & 0.80 & 0.85 & 0.78 & 0.87 & 0.87 & 0.84 & \textbf{0.84} \\
& & LPIPS $\downarrow$ & 0.21 & 0.12 & 0.38 & 0.19 & 0.30 & 0.25 & \textbf{0.25} \\ \hdashline 
\multirow{3}{*}{GLORIE-SLAM} 
& \multirow{3}{*}{Every 5 Frames} & PSNR $\uparrow$ & 21.80 & 18.66 & 17.54 & 22.85 & 19.14 & 21.52 & \textbf{20.25} \\
& & SSIM $\uparrow$ & 0.83 & 0.77 & 0.77 & 0.88 & 0.81 & 0.81 & \textbf{0.80} \\
& & LPIPS $\downarrow$ & 0.30 & 0.36 & 0.42 & 0.42 & 0.47 & 0.32 & \textbf{0.38} \\ \midrule
\multicolumn{3}{c}{Avg. FPS $\uparrow$} & 1.982 & 1.84 & 3.09 & 3.65 & 3.29 & 1.90 & \textbf{2.63} \\
\multicolumn{3}{c}{GPU Usage [GiB] $\downarrow$} & 18.978 & 15.37 & 15.90 & 13.22 & 16.10 & 15.69 & \textbf{15.97} \\
\multicolumn{3}{c}{Number of Gaussians (1000x)} & 2668 & 1695 & 1280 & 787 & 642 & 1690 & \textbf{1460} \\ \bottomrule
\end{tabular}
}
\caption{Full Evaluation on ScanNet \protect\cite{dai2017scannet}.}
\end{table*}
\clearpage

\subsection{Full Evaluation Results}
In Tables S2, S3, and S4, we present comprehensive scene-by-scene results across all commonly used evaluation metrics for the Replica, TUM-RGBD, and ScanNet datasets. The results for keyframes in the image rendering evaluation can be found in the experiment section, and Tables S2, S3, and S4 provide results for every 5 frames. Every-5-frame rendering reflects the generalization performance of the 3D model, showing superior performance compared to \cite{zhang2024glorie}. It is also worth noting that there is not a significant performance difference compared to keyframes. This highlights that our approach demonstrates excellent generalization capabilities even for unseen views. Additionally, even with a large number of Gaussians, our method ensures real-time performance and can be effectively executed on standard GPUs, not just high-end server-grade hardware

\subsection{Impact of the Number of Novel Views}
\begin{table}[ht]
\centering
\renewcommand{\arraystretch}{1.5} 
\setlength{\tabcolsep}{8pt} 
\resizebox{0.48\textwidth}{!}{
\begin{tabular}{cccccc}
\toprule
\textbf{Method} & \textbf{Metric} & \textbf{0} & \textbf{100(Ours)} & \textbf{150} \\ \midrule
\multirow{4}{*}{Keyframes} 
& PSNR $\uparrow$ & 38.871 & 39.283 & 39.310 \\
& SSIM $\uparrow$ & 0.974 & 0.976 & 0.976 \\
& LPIPS $\downarrow$ & 0.032 & 0.029 & 0.028 \\
& Depth L1 $\downarrow$ & 0.056 & 0.050 & 0.058 \\ \hdashline 
\multirow{4}{*}{Every 5 Frames} 
& PSNR $\uparrow$ & 36.475 & 37.847 & 38.002 \\
& SSIM $\uparrow$ & 0.968 & 0.971 & 0.964 \\
& LPIPS $\downarrow$ & 0.033 & 0.031 & 0.034 \\
& Depth L1 $\downarrow$ & 0.058 & 0.059 & 0.059 \\ \midrule
\multicolumn{2}{c}{FPS} & 12.07 & 9.180 & 8.54 \\ \bottomrule
\end{tabular}
}

\caption{Impact of the number of novel views}
\end{table}
We evaluated the effect of varying the number of frames used in the NVS training process following global DBA (see Table S5). Increasing the number of novel views consistently improved rendering quality but led to a decline in reconstruction quality. The enhancement in rendering quality can be attributed to the increased diversity of viewpoints, which contributes to better scene reconstruction. However, the absence of depth information in novel views resulted in a reduction in reconstruction quality. Moreover, a significant decrease in FPS was observed as the number of views increased, which presents a limitation. To achieve an optimal trade-off between real-time performance and accuracy, we selected 100 frames, as it provided the best balance for SLAM applications.

\subsection{Impact of Novel View Selection (NVS)}
\begin{figure}[ht]
    \centering
    \includegraphics[width=0.48\textwidth]{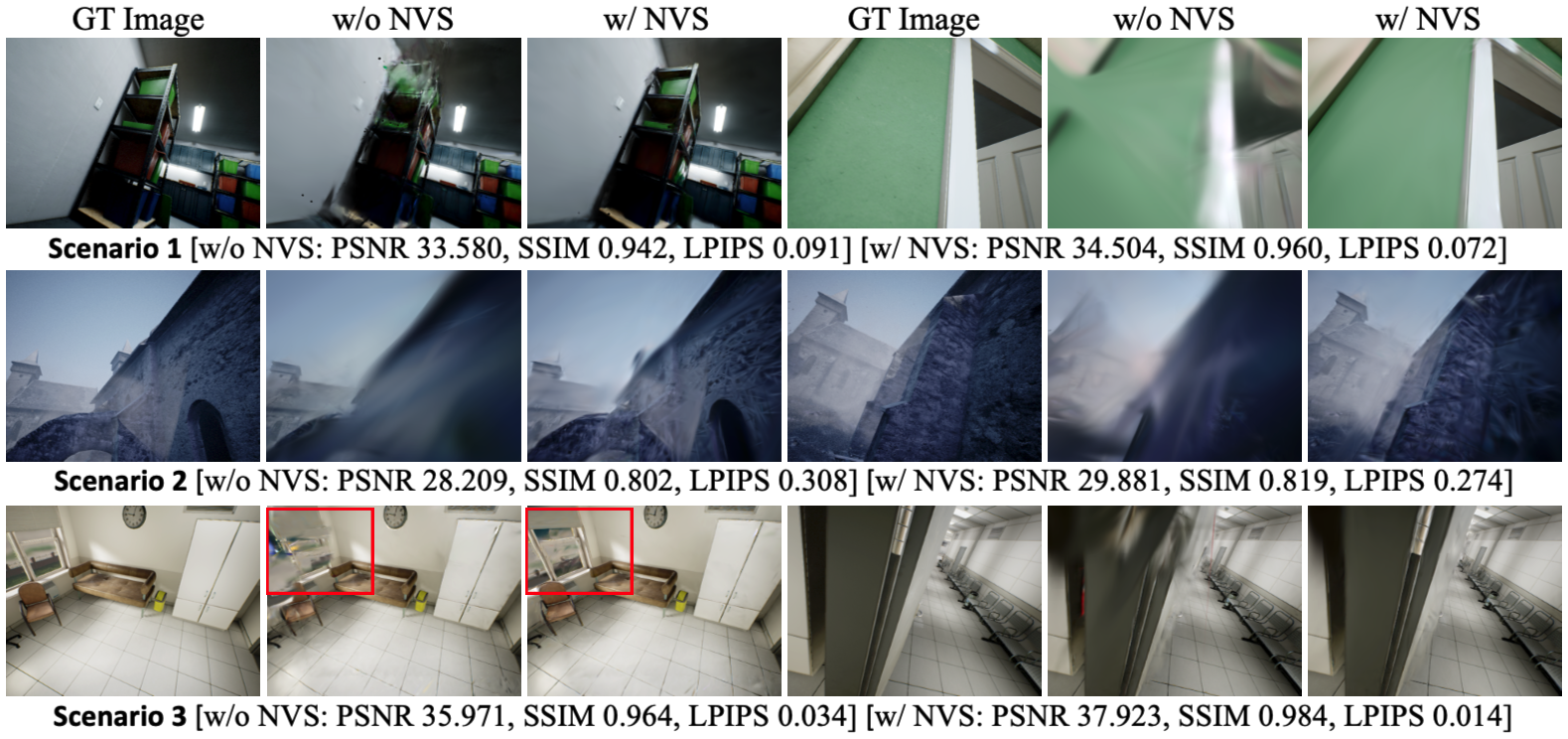}
    \caption{Comparison of results with and without NVS on the TartanAir dataset}
\end{figure}
To evaluate the effectiveness of Novel View Selection (NVS), we conducted additional experiments on the TartanAir dataset \cite{wang2020tartanair}, which contains many challenging scenes such as occlusions, motion blur, and low-texture regions. As shown in Fig. S1, applying NVS significantly improves modeling completeness and performance in these difficult cases. This demonstrates that NVS effectively handles challenging viewing conditions and enhances robustness in complex environments.

\subsection{Final Refinement Iterations}
\begin{table}[!ht]
\centering
\renewcommand{\arraystretch}{1.5} 
\setlength{\tabcolsep}{8pt} 
\resizebox{0.48\textwidth}{!}{
\begin{tabular}{cccccc}
\toprule
\textbf{Method} & \textbf{Metric} & \textbf{2K(Ours)} & \textbf{5K} & \textbf{10K} & \textbf{26K} \\ \midrule
\multirow{3}{*}{Ours} 
& PSNR $\uparrow$ & 39.28 & 40.19 & 41.18 & 42.39 \\
& SSIM $\uparrow$ & 0.98 & 0.98 & 0.98 & 0.99 \\
& LPIPS $\downarrow$ & 0.03 & 0.02 & 0.02 & 0.01 \\ \cmidrule{1-6}
Splat-SLAM & PSNR $\uparrow$ & 36.77 & 37.80 & 38.41 & 38.95 \\ \bottomrule
\end{tabular}
}
\caption{Impact of the number of final refinement iterations. Interestingly, our method, using only 2k iterations for refinement, outperforms Splat-SLAM \protect\cite{sandstrom2024splat}, which uses 26k refinement iterations}
\end{table}

After processing all sequences, we perform a final refinement inspired by the approaches in \cite{matsuki2024gaussian}, \cite{sandstrom2024splat}, and \cite{zhang2024glorie}, incorporating both color loss and geometric depth loss, rather than relying solely on color loss. As shown in Table S6, we analyze the impact of varying the number of iterations for the final refinement (2K, 5K, 10K, and 26K iterations). While rendering accuracy improves with more iterations, geometric accuracy declines. This decline is attributed to continued training on regions with low confidence in the MVS, which negatively impacts depth consistency. To strike a balance between rendering quality, geometric accuracy, and computational efficiency, we select 2K iterations as it provides the optimal trade-off.

\subsection{Disparity Map Utilization in the MVS Module}

\begin{figure}[ht]  
    \centering
    \includegraphics[width=0.48\textwidth]{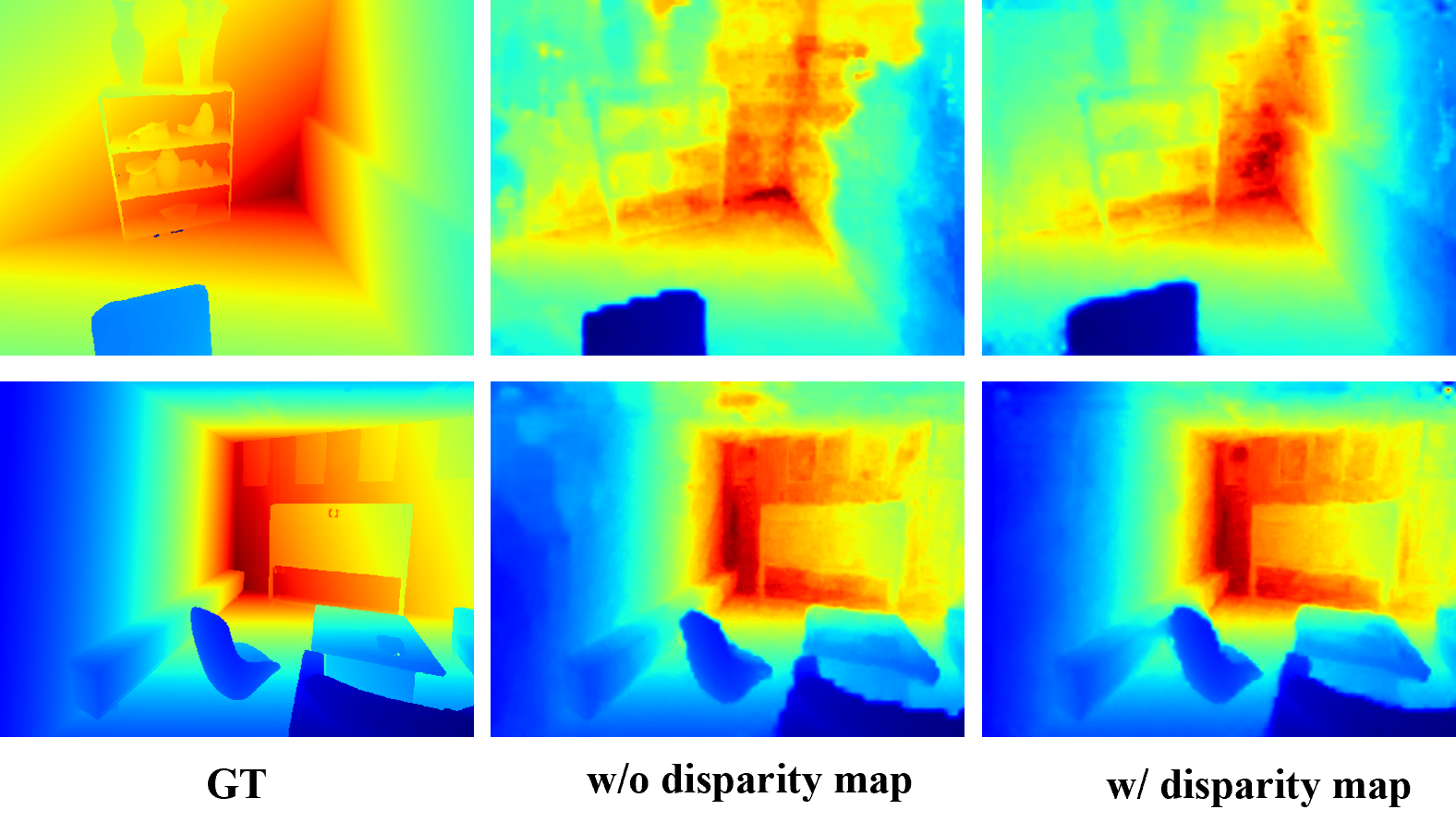}
    \caption{Effectiveness of using disparity map in the MVS module. We visualize output depths on the Replica dataset.}
\end{figure}

\begin{table}[ht]
    \centering
    \renewcommand{\arraystretch}{1.5} 
    \setlength{\tabcolsep}{8pt} 
    \resizebox{0.48\textwidth}{!}{
    {\huge
    \begin{tabular}{cccccc}
    \toprule
    \textbf{Method} & \textbf{PSNR $\uparrow$} & \textbf{SSIM $\uparrow$} & \textbf{LPIPS $\downarrow$} & \textbf{Depth L1 $\downarrow$} & \textbf{FPS $\uparrow$} \\ \midrule
    w/o disparity & 38.864 & 0.975 & 0.031 & 0.067 & 9.556 \\
    w/ disparity (Ours) & 39.283 & 0.976 & 0.029 & 0.050 & 9.180 \\ \bottomrule
    \end{tabular}
    }
    }
    \caption{Effectiveness of using disparity map in the MVS module. We report the rendering performance with and without the use of disparity maps.}
\end{table}
Instead of simply using a Multi-View Stereo (MVS) model, we integrated the disparity map estimated from DROID-SLAM into a lower layer of the MVS model (Appendix A.2). As shown in Table S7 and Fig. S2, this approach significantly enhances both reconstruction quality (Fig. S2) and rendering quality (Table S7).

\subsection{Comparison of Other View Selection}
\begin{table}[ht]
    \centering
    \resizebox{0.48\textwidth}{!}{
    \begin{tabular}{ccccc}
    \toprule
    \textbf{Method} &\textbf{PSNR $\uparrow$} & \textbf{SSIM $\uparrow$} & \textbf{LPIPS $\downarrow$} & \textbf{Depth L1 $\downarrow$} \\ \midrule
    Random & 43.288 & 0.985 & 0.020 & 0.044 \\
    Quality(PSNR) & 43.352 & 0.987 & 0.017 & 0.042 \\
    Uncertainty(Ours) & \textbf{43.932} & \textbf{0.988} & \textbf{0.016} & \textbf{0.034} \\ \bottomrule
    \end{tabular}
    }
    \caption{Ablation Study of NVS on "office0" in the Replica dataset \protect\cite{straub2019replica}}
\end{table}
We conducted experiments comparing the method of randomly selecting novel views, the PSNR-based selection method \cite{jin2024gs}, and our approach. As shown in Table S8, the results indicate that our method outperforms the others in both rendering and reconstruction quality, thereby demonstrating the effectiveness of our uncertainty-based approach.

\subsection{Efficiency Evaluation}
\begin{table}[ht]
    \centering
    \resizebox{0.48\textwidth}{!}{
    \begin{tabular}{ccccccc}
    \toprule 
    \multirow{2}{*}{Method} & \multirow{2}{*}{PSNR $\uparrow$}  & \multirow{2}{*}{SSIM $\uparrow$} & \multirow{2}{*}{LPIPS $\downarrow$} & GPU Usage & \multirow{2}{*}{Avg. FPS $\uparrow$} \\
    & & & & [GIB] $\downarrow$ & \\
    \midrule
    MonoGS       & 23.53  & 0.75 & 0.33 & 14.62 & 0.32 \\
    Splat-SLAM   & 32.25  & 0.96 & 0.09 & 17.57 & 1.24 \\
    Ours         & 34.88  & 0.96 & 0.04 & 16.34 & 10.77 \\
    \bottomrule
    \end{tabular}
    }
    \caption{Comparison of rendering accuracy and model efficiency (GPU usage and FPS) on the Replica ``room 0''.}
    \label{tab:method_comparison}
\end{table}

As shown in Table. S9, our method not only achieves high rendering quality but also demonstrates superior runtime efficiency compared to existing approaches. Specifically, it delivers a significantly higher frame rate than MonoGS and Splat-SLAM, while maintaining competitive GPU memory usage. Despite the additional computational cost introduced by NVS, our method still achieves a practical frame rate suitable for real-time applications. This highlights its effectiveness in balancing accuracy and efficiency under resource-constrained settings. 

\subsection{Runtime Analysis}
\begin{table}[ht]
    \centering
    \resizebox{0.48\textwidth}{!}{
\begin{tikzpicture}[baseline=(table.base)]
  \node[inner sep=0pt] (table) {%
    \begin{tabular}{ccccccc}
      \toprule
              & \multirow{2}{*}{\textbf{Tracking}} & \multirow{2}{*}{\textbf{MVS}}   & \multirow{2}{*}{\textbf{GBA}}   & \multirow{2}{*}{\textbf{NVS}}   & \textbf{3DGS}                   & \textbf{3DGS} \\
              &                           &                        &                        &                        & \textbf{Mapping}                & \textbf{Opt}. \\ \bottomrule
      \addlinespace[2pt]   
      Runtime & \multirow{2}{*}{0.048}    & \multirow{2}{*}{0.206} & \multirow{2}{*}{4.620} & \multirow{2}{*}{3.500} & \multirow{2}{*}{0.014} & 0.251 \\
      (sec)   &                           &                        &                        &                        &                        & (per frame) \\ \midrule
      Invocation & \multirow{2}{*}{0.050} & \multirow{2}{*}{1.010} & \multicolumn{2}{c}{\multirow{2}{*}{33.344}}     & \multicolumn{2}{c}{Consistent} \\
      Interval (sec) &                    &                        &                        &                        & \multicolumn{2}{c}{Optimization} \\
      \bottomrule
      \end{tabular}
      };
      \draw[line width=1pt] 
        ($(table.north west)+(2.3cm,-0.01cm)$) -- 
        ($(table.north west)+(2.3cm,-2.85cm)$);
      \draw[line width=1pt] 
        ($(table.north west)+(4.1cm,-0.01cm)$) -- 
        ($(table.north west)+(4.1cm,-2.85cm)$);
      \draw[line width=1pt] 
        ($(table.north west)+(5.3cm,-0.01cm)$) -- 
        ($(table.north west)+(5.3cm,-2.85cm)$);
      \draw[dashed, line width=1pt]  
        ($(table.north west)+(6.5cm,-0.01cm)$) --
        ($(table.north west)+(6.5cm,-1.85cm)$) ;
      \draw[line width=1pt] 
        ($(table.north west)+(7.7cm,-0.01cm)$) -- 
        ($(table.north west)+(7.7cm,-2.85cm)$);
      \draw[dashed, line width=1pt] 
        ($(table.north west)+(9.5cm,-0.01cm)$) --
        ($(table.north west)+(9.5cm,-1.85cm)$) ;
    \end{tikzpicture}
    }
    \caption{We report the average runtime and invocation interval of each module on the Replica ``office 0''. Modules separated by bold vertical lines run independently in parallel, while those with dashed lines are interdependent. As each module’s runtime is shorter than its invocation interval, the system avoids bottlenecks.}
\end{table}

Table. S10 summarizes the average runtime and invocation interval of each module in our system on the Replica ``office 0''. As shown in the table, all modules operate within their respective invocation intervals without exceeding them. This indicates that no single component becomes a performance bottleneck, enabling smooth and stable real-time operation.

Notably, the 3DGS optimization runs continuously and independently of the frontend modules such as tracking or MVS, ensuring consistent mapping quality without interrupting the overall pipeline. This design guarantees both efficiency and robustness, making our system suitable for long-term operation in complex environments.

\subsection{Additional Experiments}

\begin{figure*}[t]
    \centering
    \begin{minipage}{\textwidth}
        \centering
        \includegraphics[width=0.80\textwidth]{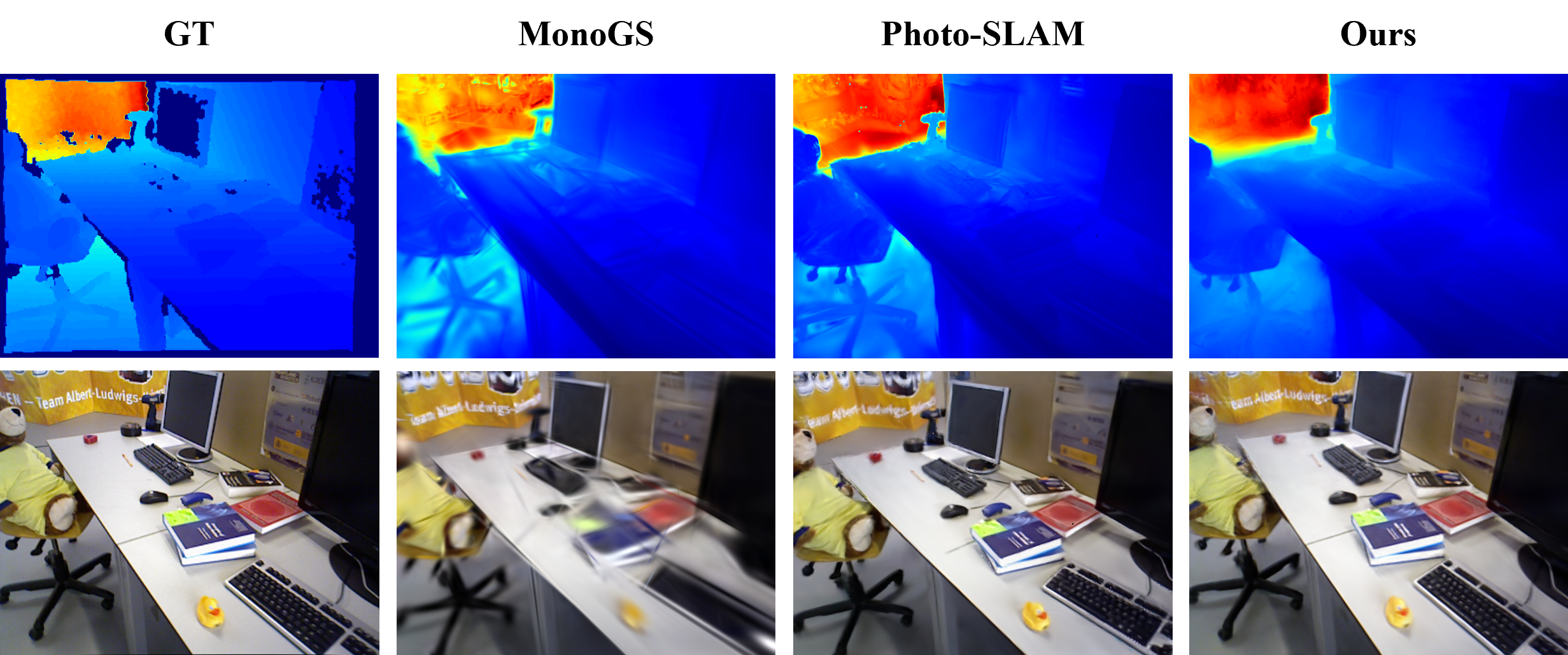}
    \end{minipage} \vspace{-4mm}  
    
    \begin{minipage}{\textwidth}
        \begin{table}[H]
            \renewcommand{\arraystretch}{1.5} 
            \setlength{\tabcolsep}{40pt} 
            \begin{tabular}{clll}
            Depth L1 [cm] $\downarrow$ & 27.00 & 38.6 & 22.56 \\ \hline
            PSNR $\uparrow$ & 19.03 & 19.51 & 28.10 \\
            \end{tabular}
        \end{table}
    \end{minipage}
    \caption{Our system delivers superior scene reconstruction and rendering quality by utilizing enhanced depth estimation techniques and an optimized Gaussian learning approach. The results are averaged across all keyframes, with the experimental scene derived from the TUM-RGBD \protect\cite{sturm2012benchmark} fr3 office dataset.}
\end{figure*}

\begin{figure*}[t]  
    \centering
        \begin{minipage}{\textwidth}
        \centering
        \includegraphics[width=0.80\textwidth]{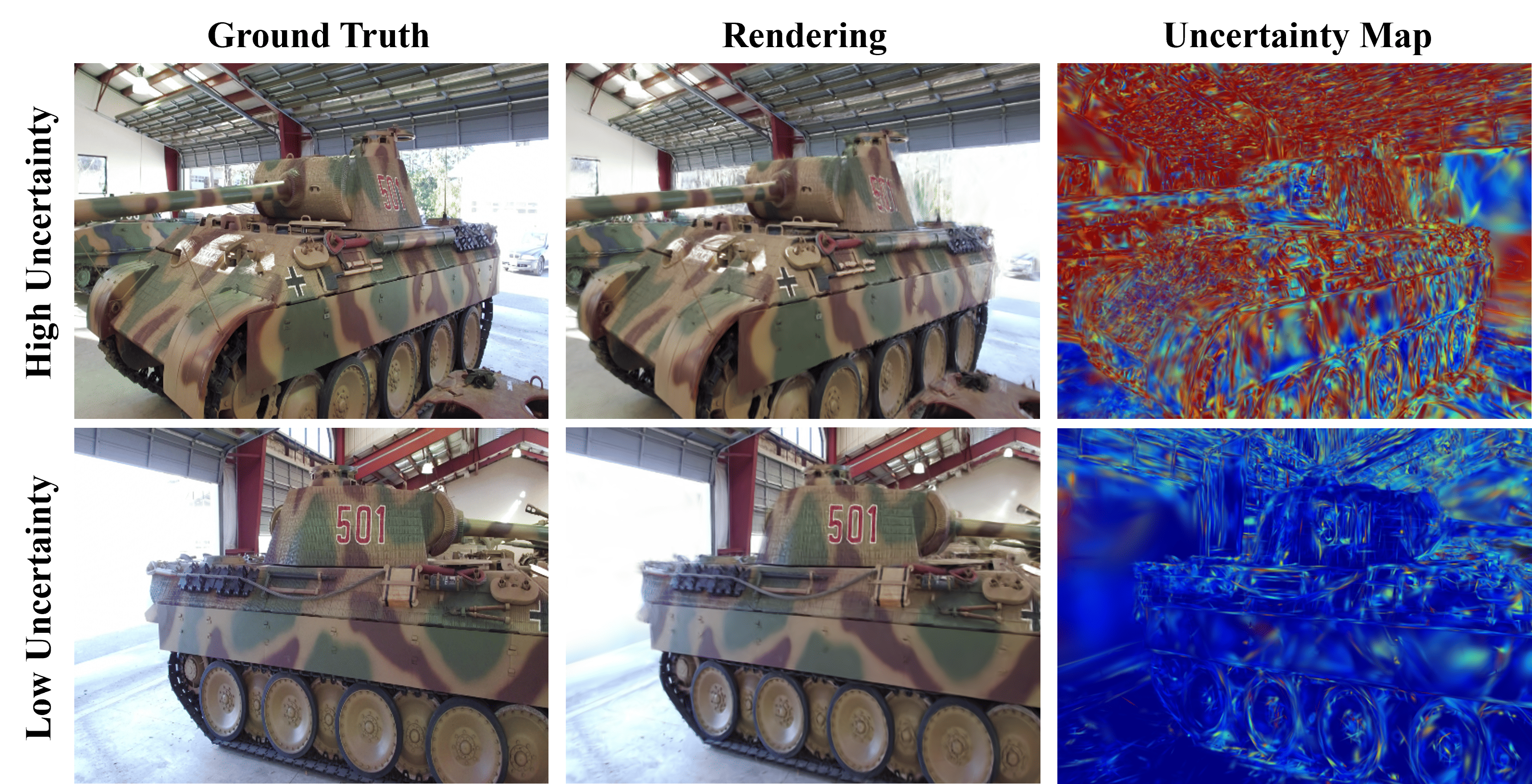}
    \end{minipage}

    \begin{minipage}{\textwidth}
        \centering
        \renewcommand{\arraystretch}{1.5} 
        \setlength{\tabcolsep}{20pt} 
    \begin{tabular}{ccccc}
        \toprule[1.5pt]
        & PSNR $\uparrow$ & SSIM $\uparrow$ & LPIPS $\downarrow$ & Uncertainty Value $\downarrow$ \\ \hline
        High Uncertainty & 19.609 & 0.632 & 0.193 & 1236.9 \\
        Low Uncertainty & 28.787 & 0.899 & 0.132 & 446.22 \\ \bottomrule[1.5pt]
    \end{tabular}
    \end{minipage}
    
    \caption{High Uncertainty vs. Low Uncertainty. We visualize the ground-truth, rendered images, and corresponding uncertainty maps for high (top) and low (bottom) uncertainty views selected by our NVS module on scene "Panther" of the Tanks \& Temples dataset \protect\cite{knapitsch2017tanks}. As observed in the third column, the high uncertainty view contains many unexplored regions (denoted in red), while the low uncertainty view is already well-reconstructed (denoted by extensive blue regions). We further demonstrate the effectiveness of our proposed NVS through quantitative metrics, as shown in the table. The results indicate that the proposed uncertainty modeling method is highly accurate and plays a crucial role in our system.}
    
\end{figure*}

\begin{figure*}[t]  
    \centering
    \includegraphics[width=0.8\textwidth]{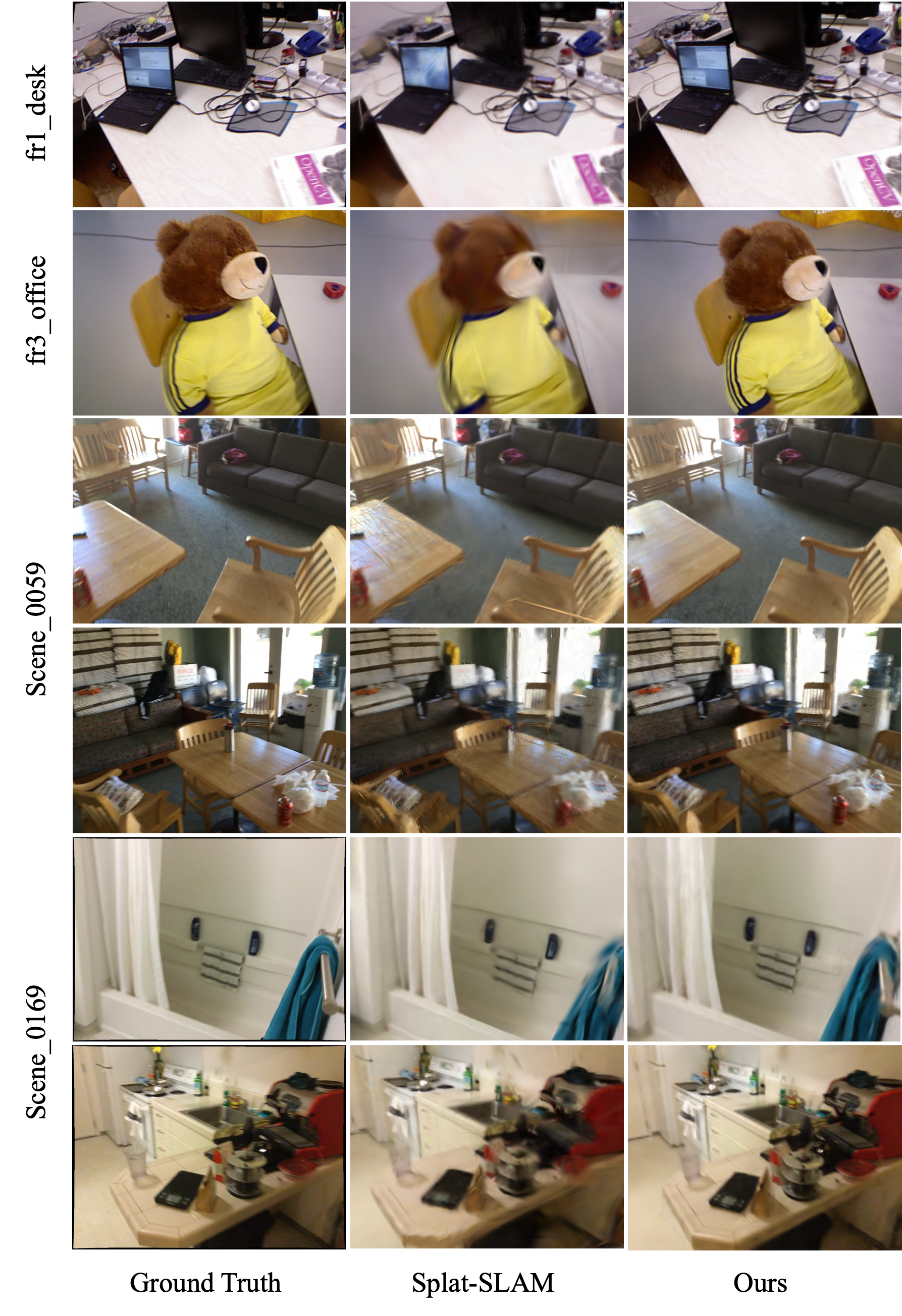}
    \caption{Qualitative results on Replica \protect\cite{straub2019replica}, TUM-RGBD \protect\cite{sturm2012benchmark} and Scannet \protect\cite{dai2017scannet}. Our method surpasses other state-of-the-art models, such as Splat-SLAM \protect\cite{sandstrom2024splat}, and occasionally produces rendering outputs that appear even more realistic than the ground truth.}
\end{figure*}

\clearpage

\begin{figure*}[t]  
    \centering
    \includegraphics[width=1\textwidth]{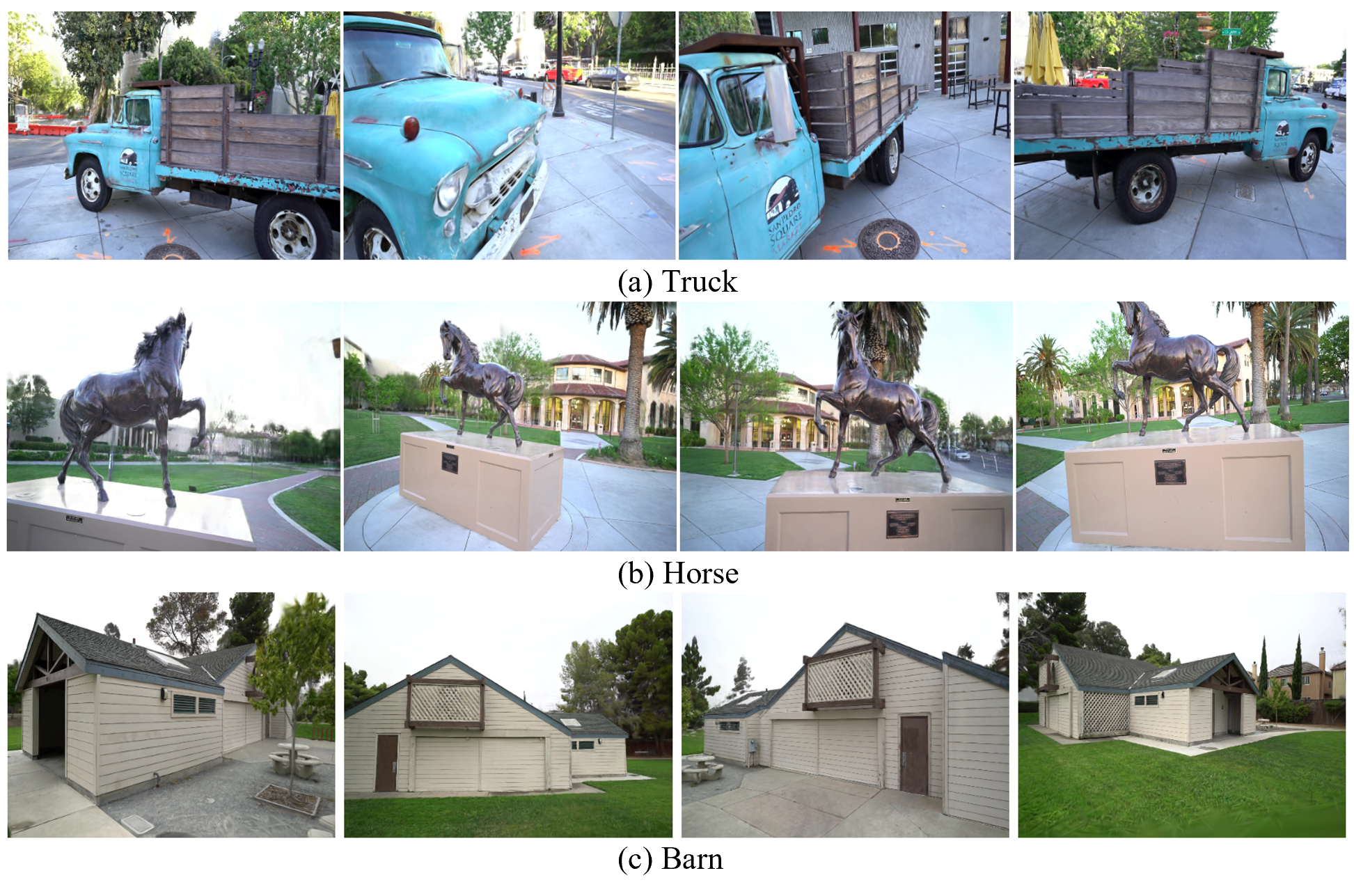}
    \caption{Qualitative results on Tanks and Temples \protect\cite{knapitsch2017tanks}.}
\end{figure*}

\begin{figure*}[t]  
    \centering
    \includegraphics[width=1\textwidth]{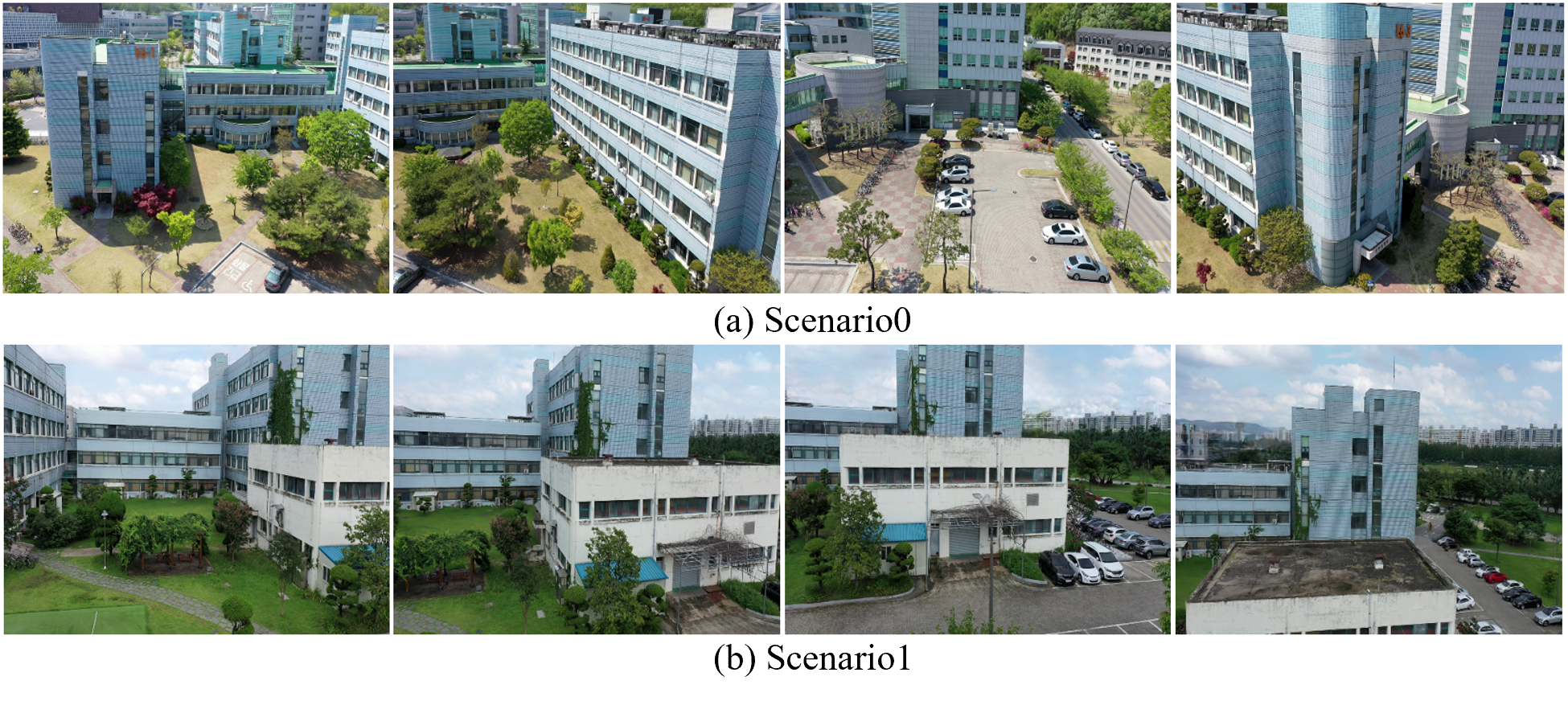}
    \caption{Qualitative results on Aerial Dataset \protect\cite{song2021view}.}
\end{figure*}

\clearpage


\bibliographystyle{named}
\bibliography{ijcai25}

\begin{thebibliography}{}

\bibitem[\protect\citeauthoryear{Bhat \bgroup \em et al.\egroup }{2023}]{bhat2023zoedepth}
Shariq~Farooq Bhat, Reiner Birkl, Diana Wofk, Peter Wonka, and Matthias M{\"u}ller.
\newblock Zoedepth: Zero-shot transfer by combining relative and metric depth.
\newblock {\em arXiv preprint arXiv:2302.12288}, 2023.

\bibitem[\protect\citeauthoryear{Bloesch \bgroup \em et al.\egroup }{2018}]{bloesch2018codeslam}
Michael Bloesch, Jan Czarnowski, Ronald Clark, Stefan Leutenegger, and Andrew~J Davison.
\newblock Codeslam—learning a compact, optimisable representation for dense visual slam.
\newblock In {\em Proceedings of the IEEE conference on computer vision and pattern recognition}, pages 2560--2568, 2018.

\bibitem[\protect\citeauthoryear{Cao \bgroup \em et al.\egroup }{2022}]{cao2022mvsformer}
Chenjie Cao, Xinlin Ren, and Yanwei Fu.
\newblock Mvsformer: Multi-view stereo by learning robust image features and temperature-based depth.
\newblock {\em arXiv preprint arXiv:2208.02541}, 2022.

\bibitem[\protect\citeauthoryear{Dai \bgroup \em et al.\egroup }{2017}]{dai2017scannet}
Angela Dai, Angel~X Chang, Manolis Savva, Maciej Halber, Thomas Funkhouser, and Matthias Nie{\ss}ner.
\newblock Scannet: Richly-annotated 3d reconstructions of indoor scenes.
\newblock In {\em Proceedings of the IEEE conference on computer vision and pattern recognition}, pages 5828--5839, 2017.

\bibitem[\protect\citeauthoryear{Eftekhar \bgroup \em et al.\egroup }{2021}]{eftekhar2021omnidata}
Ainaz Eftekhar, Alexander Sax, Jitendra Malik, and Amir Zamir.
\newblock Omnidata: A scalable pipeline for making multi-task mid-level vision datasets from 3d scans.
\newblock In {\em Proceedings of the IEEE/CVF International Conference on Computer Vision}, pages 10786--10796, 2021.

\bibitem[\protect\citeauthoryear{Engel \bgroup \em et al.\egroup }{2017}]{engel2017direct}
Jakob Engel, Vladlen Koltun, and Daniel Cremers.
\newblock Direct sparse odometry.
\newblock {\em IEEE transactions on pattern analysis and machine intelligence}, 40(3):611--625, 2017.

\bibitem[\protect\citeauthoryear{Feng \bgroup \em et al.\egroup }{2024}]{feng2024naruto}
Ziyue Feng, Huangying Zhan, Zheng Chen, Qingan Yan, Xiangyu Xu, Changjiang Cai, Bing Li, Qilun Zhu, and Yi~Xu.
\newblock Naruto: Neural active reconstruction from uncertain target observations.
\newblock In {\em Proceedings of the IEEE/CVF Conference on Computer Vision and Pattern Recognition}, pages 21572--21583, 2024.

\bibitem[\protect\citeauthoryear{Forster \bgroup \em et al.\egroup }{2014}]{forster2014svo}
Christian Forster, Matia Pizzoli, and Davide Scaramuzza.
\newblock Svo: Fast semi-direct monocular visual odometry.
\newblock In {\em 2014 IEEE international conference on robotics and automation (ICRA)}, pages 15--22. IEEE, 2014.

\bibitem[\protect\citeauthoryear{Hamdi \bgroup \em et al.\egroup }{2024}]{hamdi2024ges}
Abdullah Hamdi, Luke Melas-Kyriazi, Jinjie Mai, Guocheng Qian, Ruoshi Liu, Carl Vondrick, Bernard Ghanem, and Andrea Vedaldi.
\newblock Ges: Generalized exponential splatting for efficient radiance field rendering.
\newblock In {\em Proceedings of the IEEE/CVF Conference on Computer Vision and Pattern Recognition}, pages 19812--19822, 2024.

\bibitem[\protect\citeauthoryear{Hu \bgroup \em et al.\egroup }{2024}]{hu2024mgso}
Yan~Song Hu, Nicolas Abboud, Muhammad~Qasim Ali, Adam~Srebrnjak Yang, Imad Elhajj, Daniel Asmar, Yuhao Chen, and John~S Zelek.
\newblock Mgso: Monocular real-time photometric slam with efficient 3d gaussian splatting.
\newblock {\em arXiv preprint arXiv:2409.13055}, 2024.

\bibitem[\protect\citeauthoryear{Hu \bgroup \em et al.\egroup }{2025}]{hu2025cg}
Jiarui Hu, Xianhao Chen, Boyin Feng, Guanglin Li, Liangjing Yang, Hujun Bao, Guofeng Zhang, and Zhaopeng Cui.
\newblock Cg-slam: Efficient dense rgb-d slam in a consistent uncertainty-aware 3d gaussian field.
\newblock In {\em European Conference on Computer Vision}, pages 93--112. Springer, 2025.

\bibitem[\protect\citeauthoryear{Huang \bgroup \em et al.\egroup }{2024}]{huang2024photo}
Huajian Huang, Longwei Li, Hui Cheng, and Sai-Kit Yeung.
\newblock Photo-slam: Real-time simultaneous localization and photorealistic mapping for monocular stereo and rgb-d cameras.
\newblock In {\em Proceedings of the IEEE/CVF Conference on Computer Vision and Pattern Recognition}, pages 21584--21593, 2024.

\bibitem[\protect\citeauthoryear{Jiang \bgroup \em et al.\egroup }{2025}]{jiang2025fisherrf}
Wen Jiang, Boshu Lei, and Kostas Daniilidis.
\newblock Fisherrf: Active view selection and mapping with radiance fields using fisher information.
\newblock In {\em European Conference on Computer Vision}, pages 422--440. Springer, 2025.

\bibitem[\protect\citeauthoryear{Jin \bgroup \em et al.\egroup }{2024}]{jin2024gs}
Rui Jin, Yuman Gao, Yingjian Wang, Yuze Wu, Haojian Lu, Chao Xu, and Fei Gao.
\newblock Gs-planner: A gaussian-splatting-based planning framework for active high-fidelity reconstruction.
\newblock In {\em 2024 IEEE/RSJ International Conference on Intelligent Robots and Systems (IROS)}, pages 11202--11209. IEEE, 2024.

\bibitem[\protect\citeauthoryear{Kerbl \bgroup \em et al.\egroup }{2023}]{kerbl20233d}
Bernhard Kerbl, Georgios Kopanas, Thomas Leimk{\"u}hler, and George Drettakis.
\newblock 3d gaussian splatting for real-time radiance field rendering.
\newblock {\em ACM Trans. Graph.}, 42(4):139--1, 2023.

\bibitem[\protect\citeauthoryear{Knapitsch \bgroup \em et al.\egroup }{2017}]{knapitsch2017tanks}
Arno Knapitsch, Jaesik Park, Qian-Yi Zhou, and Vladlen Koltun.
\newblock Tanks and temples: Benchmarking large-scale scene reconstruction.
\newblock {\em ACM Transactions on Graphics (ToG)}, 36(4):1--13, 2017.

\bibitem[\protect\citeauthoryear{Koestler \bgroup \em et al.\egroup }{2022}]{koestler2022tandem}
Lukas Koestler, Nan Yang, Niclas Zeller, and Daniel Cremers.
\newblock Tandem: Tracking and dense mapping in real-time using deep multi-view stereo.
\newblock In {\em Conference on Robot Learning}, pages 34--45. PMLR, 2022.

\bibitem[\protect\citeauthoryear{Lee \bgroup \em et al.\egroup }{2024}]{lee2024mvs}
Byeonggwon Lee, Junkyu Park, Khang~Truong Giang, Sungho Jo, and Soohwan Song.
\newblock Mvs-gs: High-quality 3d gaussian splatting mapping via online multi-view stereo.
\newblock {\em arXiv preprint arXiv:2412.19130}, 2024.

\bibitem[\protect\citeauthoryear{Ma \bgroup \em et al.\egroup }{2024}]{ma2024fastscene}
Yikun Ma, Dandan Zhan, and Zhi Jin.
\newblock Fastscene: Text-driven fast 3d indoor scene generation via panoramic gaussian splatting.
\newblock In {\em IJCAI}, 2024.

\bibitem[\protect\citeauthoryear{Matsuki \bgroup \em et al.\egroup }{2024}]{matsuki2024gaussian}
Hidenobu Matsuki, Riku Murai, Paul~HJ Kelly, and Andrew~J Davison.
\newblock Gaussian splatting slam.
\newblock In {\em Proceedings of the IEEE/CVF Conference on Computer Vision and Pattern Recognition}, pages 18039--18048, 2024.

\bibitem[\protect\citeauthoryear{Mildenhall \bgroup \em et al.\egroup }{2021}]{mildenhall2021nerf}
Ben Mildenhall, Pratul~P Srinivasan, Matthew Tancik, Jonathan~T Barron, Ravi Ramamoorthi, and Ren Ng.
\newblock Nerf: Representing scenes as neural radiance fields for view synthesis.
\newblock {\em Communications of the ACM}, 65(1):99--106, 2021.

\bibitem[\protect\citeauthoryear{Mur-Artal \bgroup \em et al.\egroup }{2015}]{mur2015orb}
Raul Mur-Artal, Jose Maria~Martinez Montiel, and Juan~D Tardos.
\newblock Orb-slam: a versatile and accurate monocular slam system.
\newblock {\em IEEE transactions on robotics}, 31(5):1147--1163, 2015.

\bibitem[\protect\citeauthoryear{Newcombe \bgroup \em et al.\egroup }{2011}]{newcombe2011dtam}
Richard~A Newcombe, Steven~J Lovegrove, and Andrew~J Davison.
\newblock Dtam: Dense tracking and mapping in real-time.
\newblock In {\em 2011 international conference on computer vision}, pages 2320--2327. IEEE, 2011.

\bibitem[\protect\citeauthoryear{Peng \bgroup \em et al.\egroup }{2024}]{peng2024q}
Chensheng Peng, Chenfeng Xu, Yue Wang, Mingyu Ding, Heng Yang, Masayoshi Tomizuka, Kurt Keutzer, Marco Pavone, and Wei Zhan.
\newblock Q-slam: Quadric representations for monocular slam.
\newblock {\em arXiv preprint arXiv:2403.08125}, 2024.

\bibitem[\protect\citeauthoryear{Ran \bgroup \em et al.\egroup }{2023}]{ran2023neurar}
Yunlong Ran, Jing Zeng, Shibo He, Jiming Chen, Lincheng Li, Yingfeng Chen, Gimhee Lee, and Qi~Ye.
\newblock Neurar: Neural uncertainty for autonomous 3d reconstruction with implicit neural representations.
\newblock {\em IEEE Robotics and Automation Letters}, 8(2):1125--1132, 2023.

\bibitem[\protect\citeauthoryear{Ranftl \bgroup \em et al.\egroup }{2021}]{ranftl2021vision}
Ren{\'e} Ranftl, Alexey Bochkovskiy, and Vladlen Koltun.
\newblock Vision transformers for dense prediction.
\newblock In {\em Proceedings of the IEEE/CVF international conference on computer vision}, pages 12179--12188, 2021.

\bibitem[\protect\citeauthoryear{Redmon}{2016}]{redmon2016you}
J~Redmon.
\newblock You only look once: Unified, real-time object detection.
\newblock In {\em Proceedings of the IEEE conference on computer vision and pattern recognition}, 2016.

\bibitem[\protect\citeauthoryear{Rosinol \bgroup \em et al.\egroup }{2023}]{rosinol2023nerf}
Antoni Rosinol, John~J Leonard, and Luca Carlone.
\newblock Nerf-slam: Real-time dense monocular slam with neural radiance fields.
\newblock In {\em 2023 IEEE/RSJ International Conference on Intelligent Robots and Systems (IROS)}, pages 3437--3444. IEEE, 2023.

\bibitem[\protect\citeauthoryear{Rota~Bul{\`o} \bgroup \em et al.\egroup }{2024}]{rota2024revising}
Samuel Rota~Bul{\`o}, Lorenzo Porzi, and Peter Kontschieder.
\newblock Revising densification in gaussian splatting.
\newblock In {\em European Conference on Computer Vision}, pages 347--362. Springer, 2024.

\bibitem[\protect\citeauthoryear{Sandstr{\"o}m \bgroup \em et al.\egroup }{2024}]{sandstrom2024splat}
Erik Sandstr{\"o}m, Keisuke Tateno, Michael Oechsle, Michael Niemeyer, Luc Van~Gool, Martin~R Oswald, and Federico Tombari.
\newblock Splat-slam: Globally optimized rgb-only slam with 3d gaussians.
\newblock {\em arXiv preprint arXiv:2405.16544}, 2024.

\bibitem[\protect\citeauthoryear{Scott \bgroup \em et al.\egroup }{2003}]{scott2003view}
William~R Scott, Gerhard Roth, and Jean-Fran{\c{c}}ois Rivest.
\newblock View planning for automated three-dimensional object reconstruction and inspection.
\newblock {\em ACM Computing Surveys (CSUR)}, 35(1):64--96, 2003.

\bibitem[\protect\citeauthoryear{Song \bgroup \em et al.\egroup }{2021}]{song2021view}
Soohwan Song, Daekyum Kim, and Sunghee Choi.
\newblock View path planning via online multiview stereo for 3-d modeling of large-scale structures.
\newblock {\em IEEE Transactions on Robotics}, 38(1):372--390, 2021.

\bibitem[\protect\citeauthoryear{Straub \bgroup \em et al.\egroup }{2019}]{straub2019replica}
Julian Straub, Thomas Whelan, Lingni Ma, Yufan Chen, Erik Wijmans, Simon Green, Jakob~J Engel, Raul Mur-Artal, Carl Ren, Shobhit Verma, et~al.
\newblock The replica dataset: A digital replica of indoor spaces.
\newblock {\em arXiv preprint arXiv:1906.05797}, 2019.

\bibitem[\protect\citeauthoryear{Sturm \bgroup \em et al.\egroup }{2012}]{sturm2012benchmark}
J{\"u}rgen Sturm, Nikolas Engelhard, Felix Endres, Wolfram Burgard, and Daniel Cremers.
\newblock A benchmark for the evaluation of rgb-d slam systems.
\newblock In {\em 2012 IEEE/RSJ international conference on intelligent robots and systems}, pages 573--580. IEEE, 2012.

\bibitem[\protect\citeauthoryear{Teed and Deng}{2020}]{teed2020raft}
Zachary Teed and Jia Deng.
\newblock Raft: Recurrent all-pairs field transforms for optical flow.
\newblock In {\em Computer Vision--ECCV 2020: 16th European Conference, Glasgow, UK, August 23--28, 2020, Proceedings, Part II 16}, pages 402--419. Springer, 2020.

\bibitem[\protect\citeauthoryear{Teed and Deng}{2021}]{teed2021droid}
Zachary Teed and Jia Deng.
\newblock Droid-slam: Deep visual slam for monocular, stereo, and rgb-d cameras.
\newblock {\em Advances in neural information processing systems}, 34:16558--16569, 2021.

\bibitem[\protect\citeauthoryear{Wang \bgroup \em et al.\egroup }{2004}]{1284395}
Zhou Wang, A.C. Bovik, H.R. Sheikh, and E.P. Simoncelli.
\newblock Image quality assessment: from error visibility to structural similarity.
\newblock {\em IEEE Transactions on Image Processing}, 13(4):600--612, 2004.

\bibitem[\protect\citeauthoryear{Wang \bgroup \em et al.\egroup }{2020}]{wang2020tartanair}
Wenshan Wang, Delong Zhu, Xiangwei Wang, Yaoyu Hu, Yuheng Qiu, Chen Wang, Yafei Hu, Ashish Kapoor, and Sebastian Scherer.
\newblock Tartanair: A dataset to push the limits of visual slam.
\newblock In {\em 2020 IEEE/RSJ International Conference on Intelligent Robots and Systems (IROS)}, pages 4909--4916. IEEE, 2020.

\bibitem[\protect\citeauthoryear{Yan \bgroup \em et al.\egroup }{2023}]{yan2023active}
Dongyu Yan, Jianheng Liu, Fengyu Quan, Haoyao Chen, and Mengmeng Fu.
\newblock Active implicit object reconstruction using uncertainty-guided next-best-view optimization.
\newblock {\em IEEE Robotics and Automation Letters}, 2023.

\bibitem[\protect\citeauthoryear{Yao \bgroup \em et al.\egroup }{2018}]{yao2018mvsnet}
Yao Yao, Zixin Luo, Shiwei Li, Tian Fang, and Long Quan.
\newblock Mvsnet: Depth inference for unstructured multi-view stereo.
\newblock In {\em Proceedings of the European conference on computer vision (ECCV)}, pages 767--783, 2018.

\bibitem[\protect\citeauthoryear{Zhang \bgroup \em et al.\egroup }{2023}]{zhang2023go}
Youmin Zhang, Fabio Tosi, Stefano Mattoccia, and Matteo Poggi.
\newblock Go-slam: Global optimization for consistent 3d instant reconstruction.
\newblock In {\em Proceedings of the IEEE/CVF International Conference on Computer Vision}, pages 3727--3737, 2023.

\bibitem[\protect\citeauthoryear{Zhang \bgroup \em et al.\egroup }{2024}]{zhang2024glorie}
Ganlin Zhang, Erik Sandstr{\"o}m, Youmin Zhang, Manthan Patel, Luc Van~Gool, and Martin~R Oswald.
\newblock Glorie-slam: Globally optimized rgb-only implicit encoding point cloud slam.
\newblock {\em arXiv preprint arXiv:2403.19549}, 2024.

\bibitem[\protect\citeauthoryear{Zhou \bgroup \em et al.\egroup }{2023}]{RePaint-NeRF}
Xingchen Zhou, Ying He, F~Richard Yu, Jianqiang Li, and You Li.
\newblock {RePaint-NeRF}: Nerf editting via semantic masks and diffusion models.
\newblock In {\em IJCAI}, 2023.

\bibitem[\protect\citeauthoryear{Zhu \bgroup \em et al.\egroup }{2024}]{zhu2024nicer}
Zihan Zhu, Songyou Peng, Viktor Larsson, Zhaopeng Cui, Martin~R Oswald, Andreas Geiger, and Marc Pollefeys.
\newblock Nicer-slam: Neural implicit scene encoding for rgb slam.
\newblock In {\em 2024 International Conference on 3D Vision (3DV)}, pages 42--52. IEEE, 2024.

\end{thebibliography}

\end{document}